\documentclass{article}

\usepackage{listings}

\lstdefinestyle{Python}{
	language        =   Python, 
	basicstyle      =   \zihao{-5}\ttfamily,
	numberstyle     =   \zihao{-5}\ttfamily,
	keywordstyle    =   \color{blue},
	keywordstyle    =   [2] \color{teal},
	stringstyle     =   \color{magenta},
	commentstyle    =   \color{red}\ttfamily,
	breaklines      =   true,   
	columns         =   fixed,  
	basewidth       =   0.5em,
}

\lstdefinelanguage{Code}{
  keywords={},
  ndkeywords={class, export, boolean, throw, implements, import, this, def, self, INFO},
  sensitive=false,
  comment=[l]{//},
  morecomment=[l]{\#},
  morestring=[b]',
  morestring=[b]"
}
\usepackage{float}
\usepackage{PRIMEarxiv}
\usepackage{subfig}
\usepackage[utf8]{inputenc} 
\usepackage[T1]{fontenc}    
\usepackage{hyperref}       
\usepackage{url}            
\usepackage{booktabs}       
\usepackage{amsfonts}       
\usepackage{amssymb}
\usepackage{adjustbox}
\usepackage{nicefrac}       
\usepackage{microtype}      
\usepackage{lipsum}
\usepackage{fancyhdr}       
\usepackage{graphicx}       
\graphicspath{{imgs/}}
\usepackage{amsmath}
\usepackage{listings}       
\usepackage{xcolor}
\definecolor{LightGray}{gray}{0.9}
\usepackage{colortbl}
\usepackage{xcolor}
\usepackage{tikz}
\usepackage[edges]{forest}
\usetikzlibrary{shapes, arrows, positioning, backgrounds, fit,mindmap,shadows}
\usepackage{tcolorbox}
\usepackage[numbers, sort]{natbib}
\usepackage[figuresright]{rotating}
\usepackage{tabto}
\usepackage[section]{placeins}
\usepackage{xcolor}
\usepackage{bbding}
\usepackage{pifont}
\usepackage{utfsym}
\definecolor{bg}{gray}{0.95}
\graphicspath{{media/}}     

\usepackage{enumitem}
\setlist[itemize]{leftmargin=0.6cm}

\title{Exploring Large Language Model based Intelligent Agents: Definitions, Methods, and Prospects}
\author{
  \textbf{Yuheng Cheng}$^{1}$  \footnotemark[1]
  \quad
  \textbf{Ceyao Zhang}$^{1}$  \footnotemark[1]
  \quad
  \textbf{Zhengwen Zhang}$^{1}$
  \thanks{These authors contributed to the work equally and should be regarded as co-first authors.}
  \quad
  \\
  \textbf{Xiangrui Meng}$^1$
  \quad
  \textbf{Sirui Hong}$^2$
  \quad
  \textbf{Wenhao Li}$^1$
  \quad
  \textbf{Zihao Wang}$^3$
  \quad
  \textbf{Zekai Wang}$^4$
  \quad
  \\
  \textbf{Feng Yin}$^{1}$
  \quad  
  \textbf{Junhua Zhao}$^{1}$
  \thanks{Corresponding author}
  \quad
\textbf{Xiuqiang He}$^{5}$
  \quad
  \\[5pt]
  $^1$The Chinese University of Hong Kong, Shenzhen 
  \quad
  \\
  $^2$DeepWisdom
  \quad
  $^3$Peking University
  \quad
  $^4$Yantu.ai
  \quad
  $^5$FiT, Tencent
  \quad
}

\begin{document}
\maketitle

\begin{abstract}


Intelligent agents stand out as a potential path toward artificial general intelligence (AGI).
Thus, researchers have dedicated significant effort to diverse implementations for them.

Benefiting from recent progress in large language models (LLMs), LLM-based agents that use universal natural language as an interface exhibit robust generalization capabilities across various applications -- from serving as autonomous general-purpose task assistants to applications in coding, social, and economic domains, LLM-based agents offer extensive exploration opportunities.

This paper surveys current research to provide an in-depth overview of LLM-based intelligent agents within single-agent and multi-agent systems. It covers their definitions, research frameworks, and foundational components such as their composition, cognitive and planning methods, tool utilization, and responses to environmental feedback. We also delve into the mechanisms of deploying LLM-based agents in multi-agent systems, including multi-role collaboration, message passing, and strategies to alleviate communication issues between agents. The discussions also shed light on popular datasets and application scenarios.
We conclude by envisioning prospects for LLM-based agents, considering the evolving landscape of AI and natural language processing.
\end{abstract}

\keywords{Large Language Model \and Agent \and Multi-Agent System}

\vspace{3cm}
\noindent
\hspace*{0.2\textwidth} 
\begin{minipage}[t]{0.8\textwidth}
    We define AI as the study of agents that receive percepts from the environment and perform actions. 
    % ...
    \\
    \rule{\textwidth}{0.4pt}\\ 

    \hfill--- \textit{Artificial Intelligence: A Modern Approach}, Stuart Russell and Peter Norvig (2003). 
\end{minipage}

\newpage
\tableofcontents
\newpage

\section{Introduction}

\subsection{Intelligent Agents}
The investigation of LLM-based agents has attracted considerable attention recently. The concept of an "agent" in AI boasts a solid foundation, primarily emphasizing the distinction between agents and their environments within AI systems  \cite{russell2010artificial}. Any entity capable of perceiving its environment and taking action can be considered an agent. Agents have the autonomy to carry out tasks in diverse environments, relying on their past experiences and knowledge to make decisions that align with predefined objectives.


Generally, agents exhibit the following characteristics \cite{stone2000multiagent, wooldridge2009introduction, franklin1996agent, russell2010artificial}:

\begin{itemize}
    \item \textbf{Autonomy}: Agents independently perceive their environment, make decisions, and take actions without relying on external instructions.
    \item \textbf{Perception}: Agents are equipped with sensory capabilities that allow them to gather information about their environment through the use of sensors.
    \item \textbf{Decision-making}: Agents make decisions based on perceived information, selecting appropriate actions to achieve their goals.
    \item \textbf{Action}: Agents perform actions that alter the state of their environment.
\end{itemize}

Agents can be categorized into five types: Simple Reflex agents, Model-based Reflex agents, Goal-based agents, Utility-based agents, and Learning agents \cite{russell2010artificial}. Reinforcement Learning based agents (RL-based agents) and Large Language Model based agents (LLM-based agents) fall under the category of Learning agents.


A defining characteristic of Learning Agents is their capacity to learn and improve their behavior based on experience. These agents can enhance their decision-making processes over time by observing their environment and the results of their actions. Such improvement addresses the limitations inherent in other agent types, such as the lack of autonomous learning capabilities and difficulties in managing multi-step decision problems. These different types often depend on fixed rules or simplistic models, which can limit their adaptability and generalization abilities \cite{kaelbling1996reinforcement, prentzas2007categorizing}.

\subsection{RL-based Agents}

The primary objective of RL-based agents is to learn a policy that guides the agent to take actions in different states to maximize cumulative rewards \cite{sutton2018reinforcement}. These agents learn through trial and error, continuously adjusting their policies to optimize long-term rewards. RL-based agents have achieved considerable success \cite{mnih2015human} in domains such as gaming \cite{wang2016does}, robot control \cite{kober2013reinforcement}, and autonomous driving \cite{kiran2021deep}.

The fundamental reinforcement learning framework includes the Agent, Environment, State, Action, and Reward. The agent performs actions in the environment, and the environment responds with changes in state and rewards based on the agent's actions. The agent adjusts its policy based on the environment's feedback to attain higher cumulative rewards in future actions.

However, in recent years, certain limitations of RL-based agents have gradually emerged, representative limitations including  \cite{nguyen2020deep, mccallum1996hidden}:

\begin{itemize}
\item \textbf{Training Time}: RL algorithms often require substantial time to converge toward stable and satisfactory performance. This occurs because the agent must explore the environment, learn from its interactions, and continuously update its policy based on the observed rewards. The extended training time can present a significant drawback, particularly for large-scale and complex problems.
\item \textbf{Sample Efficiency}: RL-based agents typically must interact with the environment for many episodes before learning an effective policy. This high sample requirement can be computationally expensive and infeasible for specific applications, such as robotics or real-world scenarios where data collection is costly or time-consuming.
\item \textbf{Stability}: The learning process in RL can be unstable, particularly when using high-dimensional function approximators such as deep neural networks. This instability can lead to oscillations in performance or even divergence of the learning algorithm. This issue is exacerbated by RL-based agents often dealing with non-stationary environments, where the dynamics change as the agent's policy evolves. 
\item \textbf{Generalizability}: RL-based agents tend to be specialized in the specific task they were trained on and may not generalize effectively to new tasks or environments. This lack of generalization capability can be a significant limitation, as it requires training a new agent from scratch for each new problem. Transfer learning aims to address this issue by leveraging the knowledge acquired in one task to improve learning in a related but different task. However, developing effective transfer learning techniques for RL remains an open research challenge.
\end{itemize}

\subsection{LLM-based Agents}

Contemporary studies highlight the exceptional proficiencies of LLMs in natural language processing (NLP) domains, encompassing reasoning, general question answering, programming, and text generation \cite{zhao2023survey, bubeck2023sparks}. Nevertheless, investigations have unveiled numerous obstacles that LLMs frequently encounter while tackling pragmatic tasks \cite{chan2023chateval, devlin2019bert, reed2022generalist}:
\begin{itemize}
\item \textbf{Context Length Constraint}: LLMs frequently experience limitations in context length, with a heightened propensity for disregarding text situated in the central portion of the context compared to text at the commencement or conclusion.
\item \textbf{Protracted Knowledge Update}: LLMs necessitate considerable temporal and computational resources during each training iteration, resulting in postponed knowledge updates.
\item \textbf{Absence of Direct Tool Utilization}: LLMs cannot directly employ external instruments such as calculators, SQL executors, or code interpreters.
\end{itemize}
Incorporating agent mechanisms can facilitate the challenges above to a degree. LLM-based agents, exemplified by intelligent agents constructed upon LLMs such as GPT-4 \cite{OpenAI2023GPT4TR}, amalgamate the advantages of both LLMs and agents. Unlike other agents, LLM-based agents use LLMs for cognitive and strategic processes, encouraging smart behavior.

The merits of LLM-based agents in comparison to alternative agents encompass the following \cite{sumers2023cognitive, weng2023prompt}:

\begin{itemize}
\item \textbf{Potent Natural Language Processing and Comprehensive Knowledge}: Capitalizing on the formidable language comprehension and generation aptitudes cultivated during training on copious text data, LLMs boast substantial common sense knowledge, domain-specific expertise, and factual data. This endows LLM-based agents with the capacity to manage an array of natural language tasks.
\item \textbf{Zero-Shot or Few-Shot Learning}: LLMs have already acquired abundant knowledge and abilities during training, so LLM-based agents frequently necessitate minimal samples to excel in novel tasks. Their exceptional generalization competencies enable them to perform admirably in circumstances they have not previously encountered.
\item \textbf{Organic Human-Computer Interaction}: LLM-based agents can understand and generate natural language text, fostering interaction between human users and the intelligent agent via natural language. This augments the convenience and user-centricity of human-computer interaction.
\end{itemize}

\begin{figure}[H]
\centering
\includegraphics[width=0.65\textwidth,height=0.45\textwidth]{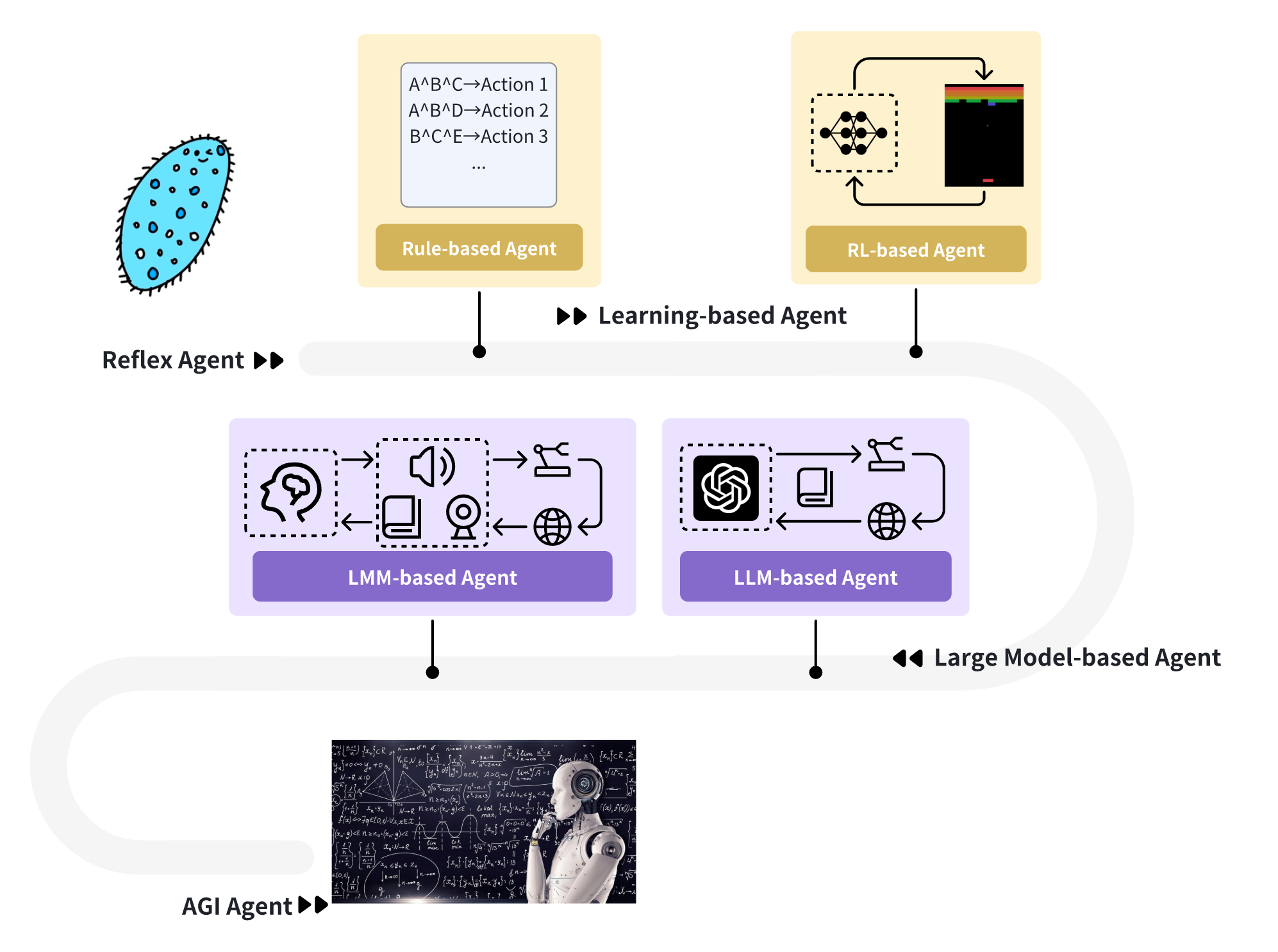}
\caption{Roadmap of Intelligent Agents Development} \label{fig_intro}
\end{figure}
By amalgamating LLMs' language comprehension and generation proficiencies with agents' decision-making and planning capabilities, LLM-based agents proffer promising resolutions to the obstacles presented by LLMs in pragmatic applications.

This paper commences with an introduction to the LLM-based agent system in Section \ref{overview}, succeeded by a synopsis of the LLM-based agent system framework in Section \ref{framework}. Section \ref{evaluation} delineates prevalent datasets and evaluation methodologies for agents. In Section \ref{applications}, we examine the employment of LLM-based agents across diverse domains, encompassing natural sciences, social sciences, engineering systems, and general domains. Conclusively, Section \ref{discussion} investigates the developmental trajectories of agents, which involve augmenting the adaptive capacity of LLM-based agents, incorporating multimodal models or large multimodal models (LMMs) to endow agents with multimodal information processing capabilities, and addressing the challenges encountered.

\section{Overview}\label{overview}
Upon scrutinizing LLM-based agents, they can be categorized into two principal classifications: Single-Agent and Multi-Agent systems. These distinct system types manifest considerable disparities in numerous facets, including application domains, memory and reconsideration mechanisms, data prerequisites, modalities, and toolsets. Subsequently, this paper delves into these agent varieties to aid readers in apprehending their singular attributes and application spheres.
\subsection{Single-Agent System}\label{Single-Agent Systems}

A single-agent system encompasses an LLM-based intelligent agent proficient in handling multiple tasks and domains, frequently denoted as an LLM-based agent. An LLM-based agent characteristically boasts extensive language comprehension, generation capacities, and multi-task generalization competencies, enabling it to execute tasks such as code generation, game exploration, and data management. Moreover, evaluation methodologies for distinct LLM-based agents vary, and the utilized tools are not standardized. An LLM-based agent may be unimodal or multimodal, depending on its design objectives. An impending table \ref{tab:single-agent} furnishes a synopsis of several contemporary LLM-based agents.

\begin{figure}[H]
\centering
\includegraphics[width=0.75\textwidth,height=0.5\textwidth]{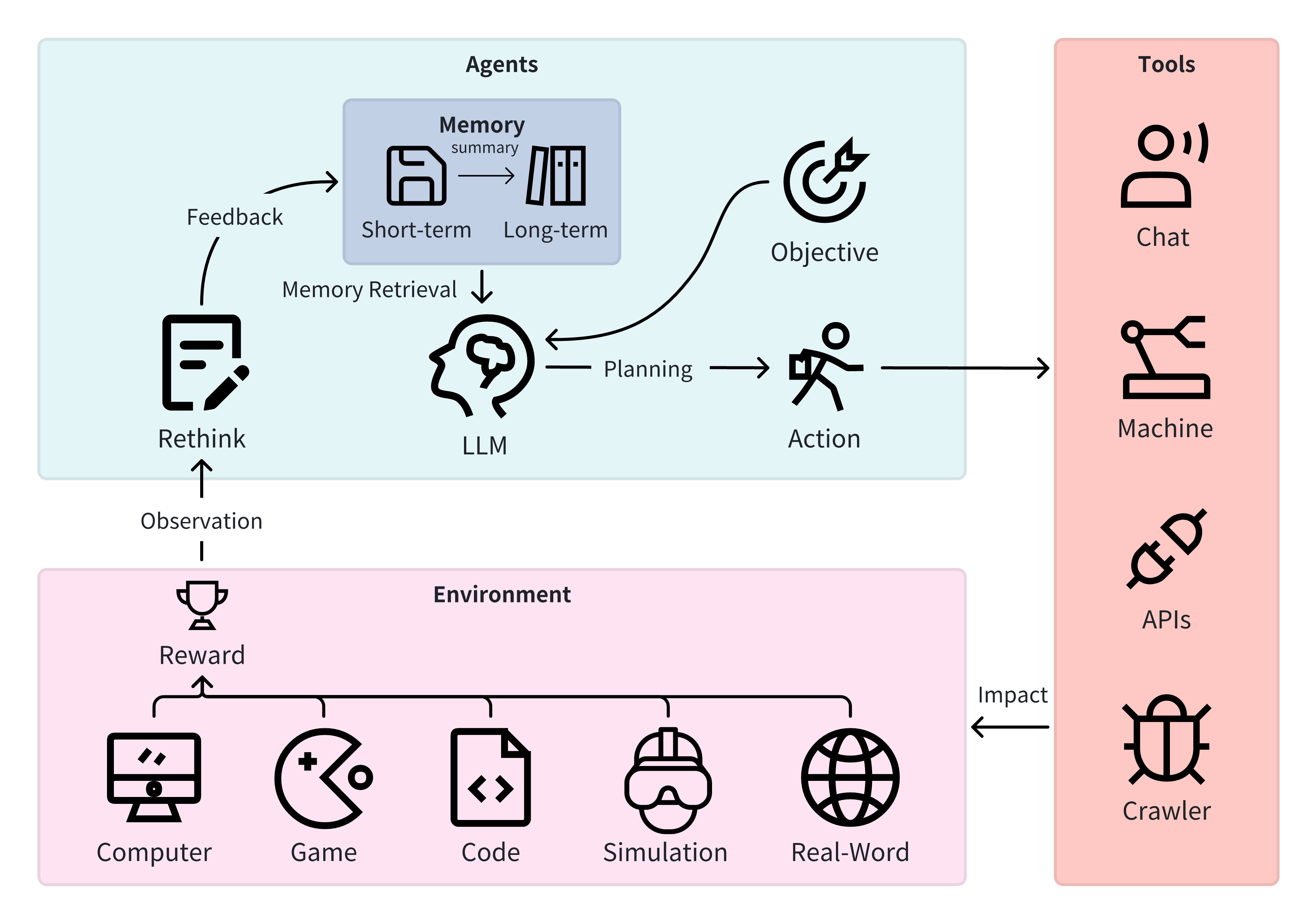}
\caption{Overview of LLM-based agents} \label{fig_agent}
\end{figure}

Each LLM-based agent $V$ can be succinctly represented as a quintuple $V = (\mathcal{L, O, M, A, R})$, wherein $\mathcal{L}$ denotes the LLM, $\mathcal{O}$ signifies the Objective, $\mathcal{M}$ embodies Memory, $\mathcal{A}$ constitutes Action, and $\mathcal{R}$ epitomizes Rethink:
\begin{itemize}
    \item \textbf{LLM}: Incorporating the LLM and the agent's configurations and proficiencies typically necessitates a prompt definition or employing a distinct domain-specific LLM. It can be posited that no supplementary training is requisite for an LLM; however, its inference parameters, such as temperature, can be dynamically adjusted. The LLM functions as the LLM-based agent's cerebral core, mandating task strategizing and decision-making predicated on current observations, historical memory, and reward information.
    \item \textbf{Objective}: The primary goal, denoted as the Objective, represents the terminal state or condition that the agent must achieve. The agent must engage in task decomposition and planning depending on the objective.
    \item \textbf{Action}: The agent possesses a repertoire of actions that can be executed, typically involving utilizing various tools, devising new tools, or transmitting messages to the environment or other agents.
    \item \textbf{Memory}: The agent's memory stores information and symbolizes the agent's current state. When the agent takes an action, the subsequent environmental feedback and reward information are recorded in the memory.
    \item \textbf{Rethink}: Upon the execution of an action, the agent is required to utilize its capacity for introspection, referred to as "Rethink," to reflect upon the preceding action and the associated environmental feedback reward. The reflective process should be integrated with the agent's memory, LLM, or other suitable models to plan and execute subsequent actions.
\end{itemize}

Regarding the external constituents of the LLM-based agent, the Environment and Tool typically comprise the following:
\begin{itemize}
    \item \textbf{Tool}: A tool refers to any instrument an agent can utilize, such as calculators, code interpreters, robotic arms, etc.
    \item \textbf{Environment}: The environment where the agent is situated significantly influences its actions. The agent can observe and interact with this environment, obtaining valuable feedback.
\end{itemize}

\subsection{Multi-Agent System}\label{Multi-Agent Systems}

Unlike a single-agent system, a multi-agent system (MAS) is a computerized system composed of multiple interacting intelligent agents \cite{hu2021decentralized}. 
Inspired by Minsky's Society of Mind (SOM)~\cite{minsky1988society} and natural language-based SOM (NLSOM)~\cite{zhuge2023mindstorms}, 
Multi-Agent Systems (MAS) design demands a heightened level of intricate coordination among various agents, particularly in their interactions and information sharing.
Each agent typically possesses specific domain expertise, making Multi-Agent systems particularly advantageous for tasks spanning multiple domains. 


\citet{decker1987distributed} outlines a four-dimensional framework for an MAS. The dimensions encompass:
1) Granularity of Agents, ranging from coarse to acceptable configurations; 
2) Heterogeneity in Agent Knowledge, comparing agents with redundant knowledge to those with specialized expertise; 
3) Mechanisms for the Distribution of Control, which can be categorized as benevolent or competitive, team-oriented or hierarchical, and may involve static or shifting role assignments; 
4) Varieties of Communication Protocols, differentiating between the blackboard and message-based systems and specifying the gradation from low-level to high-level contents.

From an application perspective, \citet{parunak1996applications} presented a taxonomy of MAS from three important characteristics:
\begin{itemize}
    \item \textbf{System Function};
    \item \textbf{System Architecture} (e.g., communication, protocols, human involvement);
    \item \textbf{Agent Architecture} (e.g., degree of heterogeneity, reactive vs. deliberative).
\end{itemize}

The main contribution of the taxonomy lies in dividing MAS into agent-level and system-level characteristics.

\citet{stone2000multiagent} classifies MAS according to two crucial dimensions: the degree of heterogeneity and the degree of communication. This classification framework yields four distinct archetypes of MAS: homogeneous non-communicating agents, heterogeneous non-communicating agents, homogeneous communicating agents, and heterogeneous communicating agents.
Incorporating approaches such as control theory and reinforcement learning is a common practice to bestow intelligence and autonomy upon these agents.

As highlighted in \citet{yang2021many}, following the breakthrough of the deep Q-learning (DQN) \cite{mnih2015human} architecture in a single-agent paradigm, 2019 observed RL-based agents expanding to multi-agent systems, signifying a burgeoning of Multi-Agent Reinforcement Learning (MARL) techniques. Within the context of MARL, \citet{hu2022marllib} offers a taxonomy to distinguish MARL algorithms by employing the subsequent four dimensions:
\begin{itemize}
    \item \textbf{Task Mode}: Cooperative-like or Competitive-like;
    \item \textbf{Agents Type}: Heterogeneous or Homogeneous;
    \item \textbf{Learning Style}: Independent Learning, Centralized Training, Decentralized Execution (CTDE), or Fully Centralized;
    \item \textbf{Knowledge Sharing}: Agent Level, Scenario Level, or Task Level.
\end{itemize}

The LLM has been flourishing since 2022. Considering the LLM-based agents in MAS, a graph $G(V, E)$ can represent the relationships among multiple LLM-based agents. Here $V$ is the set of nodes, with $V_i$ representing an LLM-based agent, and $E$ is the set of edges, with $E_{ij}$ representing the message passing and relationship between LLM-based agents $V_i$ and $V_j$. 

We propose a categorization by taking into account the following aspects:
\begin{itemize}
\item \textbf{Multi-Role Coordination}: Cooperative, Competitive, Mixed, and Hierarchical;
\item \textbf{Planning Type}: Centralized Planning Decentralized Execution (CPDE) and Decentralized Planning Decentralized Execution (DPDE).
\end{itemize}

\begin{figure}[H]
\centering
\includegraphics[width=0.75\textwidth,height=0.5\textwidth]{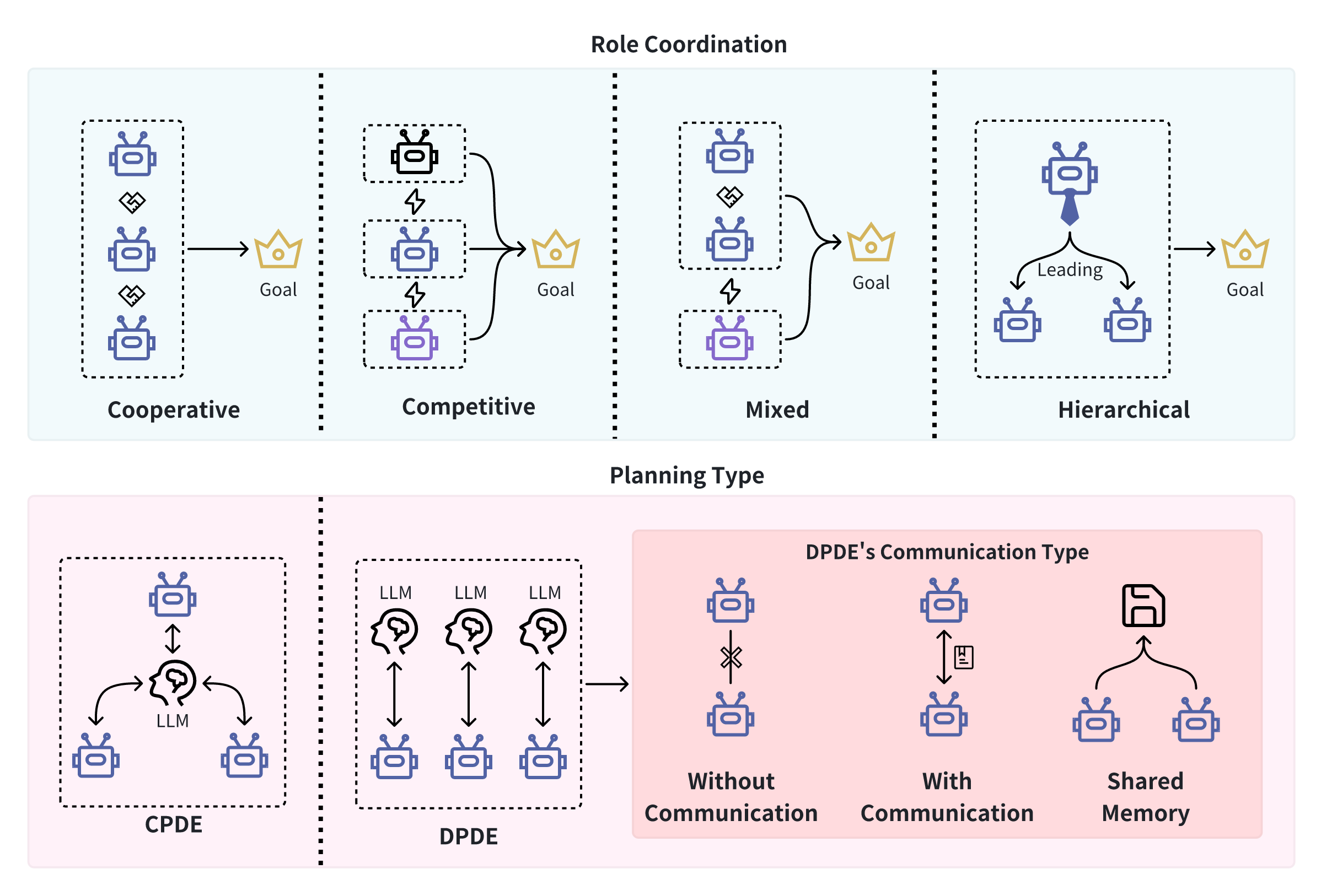}
\caption{The Relationship between LLM-based agents} \label{fig_mas}
\end{figure}

It will be listed with the detailed information of each LLM-based in the following table \ref{tab:multi-agent}.

\begin{sidewaystable}[htb]
\centering
\caption{List of LLM-based Single-Agent System. }
\label{tab:single-agent}
\renewcommand\tabcolsep{1pt}
\renewcommand\arraystretch{1.2}
\small
\begin{tabular}{rcllllllllllll}
\toprule
ID&Name&Field&Training&Environment&Data&Evaluation&Modality&Feedback&Tool&Planning&Review\\
\midrule

1&Out of One\cite{argyle2023out}&Sociology&No&Text&No& with human&Text&None&None&None&None\\
2&\citet{horton2023large}&Sociology&No&Text&No& with human&Text&None&None&None&None\\
3&\citet{park2022social}&Sociology&No&Text&No& with human&Text&None&None&None&None\\
4&Social AI School\cite{kovavc2023socialai}&Sociology&Yes&Simulation environment&Yes&Comparison available&Image to text&self-feedback&None&None&RL\\
5&$S^3$\cite{gao2023s}&Sociology&No&Text&No& with humans&Text&environmental&None&None&None\\
6&\citet{li2023you}&Sociology&No&Text&No&None&Text&None&Yes&None&None\\
7&\citet{li2023quantifying}&Sociology&No&Text&None& with human&Text&None&None&None&None\\
8&ChatLaw\cite{cui2023chatlaw}&Law&Yes&Text&Yes& with models&Text&None&None&None&None\\
9&\cite{hamilton2023blind}&Law&Yes&Text&Yes& with models&Text&None&None&None&None\\
10&ChemCrow\cite{bran2023chemcrow}&Chemistry&Yes&Experimental environment&Yes&LLM and expert assessment&Multimodal&self-feedback&Yes&ICL&ICL\\
11&ChatMOF\cite{kang2023chatmof}&Material Science&No&Experimental environment&Yes&success rate available&Multimodal&self-feedback&Yes&ICL&ICL\\
12&Mathagent\cite{swan2023math}&Mathematics&Yes&Text&Yes&None&Text&None&None&None&None\\
13&\cite{drori2022neural}&Mathematics&Yes&Text&Yes& with models&Text&None&None&None&None\\
14&IGLU\cite{mehta2023improving}&Collaborative&Yes&IGLU competition&No& with models&Text&humanfeedback&None&None&RL\\
15&GPTEngineer\cite{githubGitHubAntonOsikagptengineer}&Code&No&Code Environment&No&None&Text&self-feedback&Yes&ICL&ICL\\
16&SmolModels\cite{githubGitHubSmolaideveloper}&Code&No&Code Environment&No&None&Text&self-feedback&Yes&ICL&ICL\\
17&DemoGPT\cite{githubGitHubMelihunsalDemoGPT}&Code&Support&Code Environment&Support&None&Text&self-feedback&Yes&ICL&ICL\\
18&IELLM\cite{ogundare2023industrial}&Industrial Engineering&No&None&None&None&None&No&No&No&No\\
19&DialogueShaping\cite{zhou2023dialogue}&Game&No&Game environment&No& with agents&Text&environmental &No&ICL&RL\\
20&DECKARD\cite{nottingham2023embodied}&Game&No&Game environment&No&ablation study&Text&environmental &No&Multi-stage&RL\\
21&TaPA\cite{song2022llm}&Embodied agents&No&Visual perception environment&No& with models&Visual  &environmental &No&ICL&ICL\\
22&Voyager\cite{wang2023voyager}&Game&No&Minecraft game &No& with models&Text&environmental &No&ICL&ICL\\
23&GITM\cite{zhu2023ghost}&Game&No&Minecraft game &No& with models&Text&environmental&No&ICL&ICL\\
24&LLM4RL\cite{huenabling}&Embodied agents&Yes&Different embodied environments&No& with baseline&Multi&environmental &No&None&RL\\
25&PET\cite{wu2023plan}&Embodied agents&Yes&AlfWorld interactive environment&Yes& with models&Text&self-feedback&No&Multi-stage&None\\
26&REMEMBERER\cite{zhang2023large}&None&No&Text&No& with  models&Text&No&Yes&None&RL\\
27&UnifiedAgent\cite{di2023towards}&Embodiedagents&Yes&Robot simulation environment&Yes& ablation study&Multi&None&None&None&RL\\
28&SayCan\cite{saycan2022arxiv}&Embodied agents&No&Robot simulation environment&Yes&ablation study&Multi&No&None& External Method&RL\\
29&AIlegion\cite{githubGitHubEumemicailegion}&Universal&No&No&No&None&Multi&None&Yes&ICL&ICL\\
30&AGiXT\cite{githubGitHubJoshXTAGiXT}&Universal&No&No&No&None&Multi-modal&None&Yes&ICL&ICL\\
31&BabyAGI\cite{githubGitHubYoheinakajimababyagi}&Embodied agents&No&No&No&None&Multi&None&Yes&ICL&ICL\\
32&LoopGPT\cite{githubGitHubFarizrahman4uloopgpt}&Universal&None&None&None&None&Multi-modal&&Yes&ICL&ICL\\
33&GPTresearcher\cite{githubGitHubAssafelovicgptresearcher}&Research&None&None&None&None&Multi&No&Yes&External Method&None\\
34&SuperAGI\cite{githubGitHubTransformerOptimusSuperAGI}&Universal&None&None&None&None&Multi-modal&None&Yes&ICL&ICL\\

\bottomrule
\end{tabular}

\end{sidewaystable}

\begin{sidewaystable}[htb]

\centering
\caption{List of LLM-based Multi-Agent System. }
\label{tab:multi-agent}
\renewcommand\tabcolsep{1pt}
\renewcommand\arraystretch{1.2}
\small
\resizebox{\textwidth}{!}{%
\begin{tabular}{rcllllllllllll}
\toprule
ID&Name&Field&Training&Environment&Data&Evaluation&Modality&Tool&Planning&Review\\
\midrule

1&Generative Agents\cite{park2023generative}&Sociology&No&GameText&No&None&Text&No&ICL&reflection&Cooperative&CPDE\\
2&\citet{williams2023epidemic}&Epidemiology research&No&Text&None&None&Text&None&None&None&Cooperative&CPDE\\
3&\citet{boiko2023emergent}&Chemistry&No&Experimental environment&No&None&Multi&Yes&ICL&ICL&Hierarchical&DPDE\\
4&ChatDev\cite{qian2023communicative}&Software Development&No&Software Development&No&on dataset available&Text&Yes&ICL&ICL&Cooperative&DPDE\\
5&MetaGPT\cite{hong2023metagpt}&Code&Support&Code Environment&Support& with models&Multi&Support&ICL&ICL&Cooperative&DPDE\\
6&SCG\cite{dong2023self}&Code&No&Code Environment&No& with models&Text&Yes&ICL&ICL&Cooperative&DPDE\\
7&GPT4IA\cite{xia2023towards}&Industrial environment&Yes&Engineering environment&No&None&Multi&Yes&ICL&ICL&Cooperative&DPDE\\
8&Planner-Actor-Reporter\cite{dasgupta2023collaborating}&Embodied environments&Support&A 2D partially observable grid-world&No&task success rate&Image plus text&No&ICL&ICL&Cooperative&DPDE\\
9&ProAgent\cite{zhang2023proagent}&Universal&No&Text&No&with models&Multi&No&None&None&Hierarchical&DPDE\\
10&SAMA\cite{li2023semantically}&Game&No&Game&No&with models&GameText&No&ICL&ICL and RL&Cooperative&DPDE\\
11& Co-LLM-Agents\cite{zhang2023building}&Game&No&Game&No&with models&GameText&No&None&None&Cooperative&DPDE\\

\bottomrule
\end{tabular}
}
\end{sidewaystable}

\subsection{Agent System Template}\label{Template}

Numerous researchers propose agent and template solutions to aid future researchers and enthusiasts in developing more pertinent agents. For instance, ToolLLM \cite{qin2023toolllm} presents a comprehensive template for data construction, model training, and evaluation, fostering the development of agents with enhanced functionalities.

Various projects, such as AutoGPT \cite{githubGitHubSignificantGravitasAutoGPT}, XLang \cite{githubGitHubXlangaixlang}, LangChain \cite{githubGitHubLangchainailangchain}, MiniAGI \cite{githubGitHubXlangaixlang}, XAgent \cite{xagent2023}, OpenAgents \cite{xie2023openagents}, and WorkGPT \cite{githubGitHubTeamopenpmworkgpt}, have open-sourced their code on GitHub. These templates support diverse functionalities, encompassing distinct approaches to thinking, planning, and reviewing, and permit the integration of various models as the agent's core component. Furthermore, AgentGPT \cite{githubGitHubReworkdAgentGPT} provides features for fine-tuning models and incorporating local data into the model training process. \citet{crouse2023formally} introduces a streamlined template, utilizing Linear Temporal Logic (LTL) to facilitate the design and implementation of LLM-based agents, promoting rapid experimentation and enhancing agent performance.

Additionally, templates such as AutoGen \cite{wu2023autogen}, AgentVerse \cite{chen2023agentverse}, AutoAgents \cite{chen2023auto}, and AGENTS \cite{zhou2023agents} expedite the creation of multi-agent systems by allowing the selection and customization of roles within a multi-agent configuration, thereby simplifying the development process.

\section{LLM-based Agent System Framework}\label{framework}

\subsection{LLM-based Single Agent System}\label{framework_single_agent}

This section succinctly dissects the single-agent system into five key components: Planning, Memory, Rethinking, Environment, and Action. Each component, highlighted for its unique contribution, forms a crucial part of the unified whole, underscoring the system's intricate design and functionality.

\subsubsection{Planning}\label{framework_single_agent_planning} The planning capability defines an LLM-based agent's ability to devise action sequences based on set objectives and existing environment constraints, ensuring goal fulfillment. It's a vital feature of LLM-based agents, encompassing task analysis, potential action anticipation, optimal action selection, and the ability to tackle complex problems and tasks.
Unlike conventional and RL agents that use planning algorithms like Dijkstra \cite{noto2000method}, and POMDP \cite{cassandra1998survey} to find the best action sequence in state spaces and plan in uncertain environments, RL-based agents require learning policies \cite{kaelbling1996reinforcement}. 
LLM-based agents derive their planning capabilities primarily from the LLM. Even though LLMs primarily communicate through natural language or specific text, their internal structures and training methods bestow upon them a level of planning proficiency. Recent research trends also highlight guiding LLMs in thinking and planning as a crucial development direction.

\begin{figure*}[!ht]
\scriptsize
\begin{adjustbox}{width=\textwidth}
\begin{forest}
  for tree={
    forked edges,
    grow'=0,
    draw,
    rounded corners,
    node options={align=center},
    text width=3cm,
    s sep=5pt,
    calign=edge midpoint, 
    fill=white,
    drop shadow,
  }
  [Planning Capability of LLM-based Agents, fill=gray!45
    [In-Context Learning Methods
      [Chain of Thought (CoT)\cite{wei2022chain, fu2022complexity, zhang2022automatic, kojima2022large}]
      [Self-consistency \cite{wang2022self}]
      [Tree of Thought \cite{yao2023tree}]
      [Least-to-Most \cite{zhou2022least}]
      [Skeleton of Thought \cite{ning2023skeleton}]
      [Graph of Thought \cite{besta2023graph}]
      [Progressive Hint Prompting \cite{zheng2023progressive}]
      [Self-Refine \cite{madaan2023self}]
    ]
    [External Methods
      [LLM+P \cite{liu2023llm+}]
      [LLM-DP \cite{dagan2023dynamic}]
      [RAP \cite{hao2023reasoning}]
      [\citet{romero2023synergistic}]
      [ \citet{merkle2023context}]
    ]
    [Multi-stage Methods
      [SwiftSage \cite{lin2023swiftsage}]
      [DECKARD \cite{nottingham2023embodied}]
    ]
  ]
\end{forest}
\end{adjustbox} 
\caption{Typology of the Planning capability.}
\label{fig:planning_mindmap}
\end{figure*}
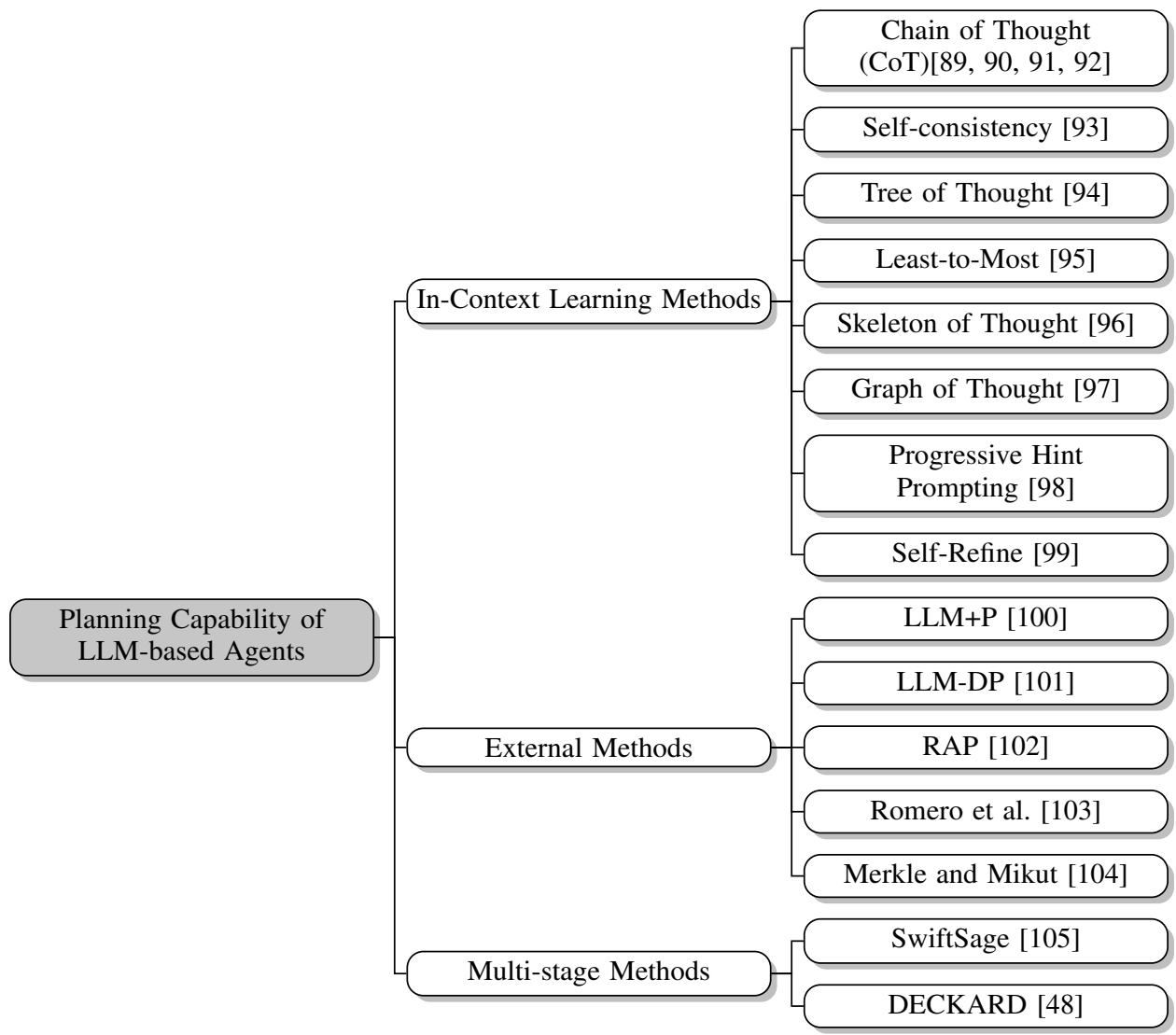

\paragraph{In-Context Learning (ICL) Methods} ICL utilizes natural language prompts, comprising task descriptions and potentially supplemented by task examples, to guide language models in problem-solving \cite{dong2022survey}. Chain of Thought (CoT), encompassing Complex CoT \cite{fu2022complexity}, Auto CoT \cite{zhang2022automatic}, and zero-shot CoT \cite{kojima2022large}, employs thought-guided prompting to systematically deconstruct intricate tasks into smaller, manageable components, thus facilitating long-term planning and deliberation. To augment CoT's efficacy, Self-consistency \cite{wang2022self} generates multiple reasoning pathways using an LLM and integrates the resulting answers, for example, by selecting the most consistent response through voting among the pathways. Tree of Thought (ToT) \cite{yao2023tree} segregates problems into several thinking stages, producing multiple concepts at each stage and forming a tree-like structure. The search process implements breadth-first or depth-first exploration and evaluates each state using classifiers or majority voting. 

To enhance CoT's generalization abilities, Least-to-Most \cite{zhou2022least} disassembles complex problems into sub-problems and sequentially addresses them. Concurrently, Skeleton of Thought (SoT) \cite{ning2023skeleton} initially directs the LLM to generate an answer's framework and subsequently complete each skeleton point through API calls or batch decoding, significantly expediting answer generation. Graph of Thought (GoT) \cite{besta2023graph} represents the information produced by the LLM as an arbitrary graph, with information units (LLM thoughts) as vertices and edges corresponding to dependencies between these vertices. Progressive Hint Prompting (PHP) \cite{zheng2023progressive} hastens guidance toward accurate answers by employing previously generated responses as prompts, thus improving the model's reasoning capabilities in problem-solving contexts. Self-Refine \cite{madaan2023selfrefine} enables the LLM to offer multifaceted feedback on its outputs and iteratively refine prior outputs based on this feedback, emulating the iterative improvement process humans may experience when generating text.

\paragraph{External Capabilities Methods} Using external capabilities methods involves employing tools, algorithms, or simulation techniques for planning purposes in computer science. LLM+P \cite{liu2023llm+} relies on classical planners for long-term planning, utilizing the Planning Domain Definition Language (PDDL) \cite{aeronautiques1998pddl} as an intermediate interface. The model translates the problem into a problem description (problem PDDL), requests the planner to generate a PDDL plan based on the "Domain PDDL," and then converts the PDDL plan back into natural language. LLM-DP \cite{dagan2023dynamic} combines LLM with symbolic planners for solving embodied tasks, leveraging LLM's understanding of the impact of actions on the environment and the planner's solution-finding efficiency.  \citet{guan2023leveraging} utilizes GPT-4 to generate PDDL, refines the PDDL with natural language feedback, and applies the extracted domain models for robust planning across various methods. RAP \cite{hao2023reasoning} framework implements conscious planning reasoning in LLM by adding a world model. It employs principled planning, specifically Monte Carlo Tree Search, for efficient exploration to generate high-reward reasoning trajectories. 

In addition to these methods, several other approaches have been proposed to enhance planning and reasoning capabilities. \citet{zhao2023large} employs LLMs as commonsense world models and applies heuristic strategies to address complex task-planning problems. \citet{romero2023synergistic} outlines a feasible approach to integrating Cognitive Architectures and LLM. \citet{merkle2023context} proposes a simulation-based method that represents heterogeneous contexts through knowledge graphs and entity embeddings and dynamically composes policies through parallel-running agents. FaR \cite{zhou2023far} combined with Theory of Mind (ToM) \cite{premack1978does} offers a framework that enables the LLM to anticipate future challenges and contemplate potential actions. LATS \cite{zhou2023language} incorporates LLM as an intelligent agent, value function, and optimizer, leveraging its potential benefits to augment planning, action, and reasoning capabilities. Think-on-Graph \cite{sun2023thinkongraph} facilitates agents in identifying the optimal planning path by executing beam search on the knowledge graph. These approaches demonstrate the versatility and potential of LLMs in various planning and reasoning tasks, paving the way for future more advanced and efficient solutions.

\paragraph{Multi-stage Methods} Multi-stage methods dissect the planning process into distinct stages, aiming to improve LLM's performance in complex reasoning and problem-solving tasks. SwiftSage \cite{lin2023swiftsage} is a framework inspired by the dual-process theory that combines the advantages of behavior cloning and guided LLMs to enhance task completion performance and efficiency. It consists of two primary modules: the SWIFT module, responsible for rapid, intuitive thinking, and the SAGE module, handling deliberative thinking. The exploration process of DECKARD \cite{nottingham2023embodied} is divided into the Dreaming and Awake stages. During the Dreaming stage, the agent utilizes an LLM to decompose the task into sub-goals. In the Awake stage, the agent learns a modular strategy for each sub-goal, verifying or rectifying assumptions based on the agent's experience. 

\vspace{.5em}

These methods enhance the model's performance in complex reasoning and problem-solving tasks. Through these methods, LLM can be guided in thinking and planning to address intricate problems and tasks.

\subsubsection{Memory} The primary function of the memory system in LLM-based agents is to preserve and regulate knowledge, experiential data, and historical information, which can be utilized for reference and modification during problem-solving and task execution processes. Additionally, the memory frequently embodies the present state of the LLM-based agent. Conventionally, the memory of such agents is documented in a textual format, enabling seamless interaction with the LLM. This paper delineates prevalent memory classifications and their associated design approaches.

\begin{figure*}[!ht]
\scriptsize
\begin{adjustbox}{width=\textwidth}
\begin{forest}
  for tree={
    forked edges,
    grow'=0,
    draw,
    rounded corners,
    node options={align=center},
    text width=3cm,
    s sep=5pt,
    calign=edge midpoint, 
    fill=white,
    drop shadow,
  }
  [Memory Capability of LLM-based Agents, fill=gray!45
    [Short-term Memory
    ]
    [Long-term Memory
      [Knowledge Graphs\cite{wang2017knowledge}]
      [Vector Databases\cite{singla2021experimental}]
      [Relational Database Queries]
      [API Calls]
    ]
    [Memory Retrieval
    ]
  ]
\end{forest}
\end{adjustbox} 
\caption{Typology of the Memory.}
\label{fig:memory_mindmap}
\end{figure*}
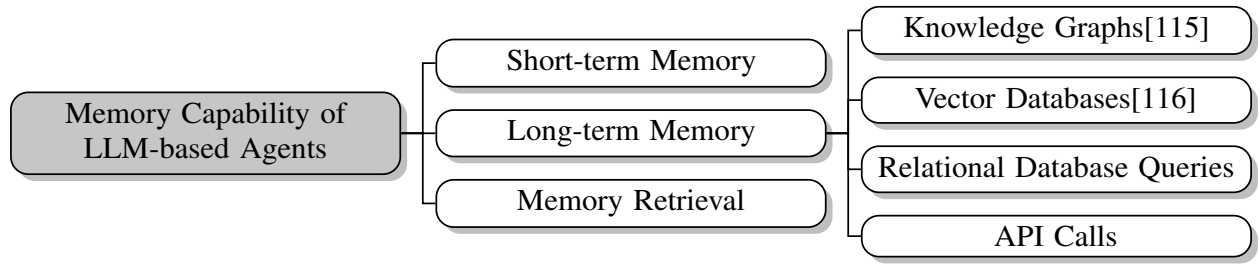

\paragraph{Short-term Memory} Short-term memory stores and manipulates a restricted quantity of transient information. Within the context of an LLM-based agent, this can be realized by amalgamating input text with contextually pertinent data related to the ongoing task, bound by the LLM's context length. As demonstrated by ChatDev \cite{qian2023communicative}, the conversation history is archived, enabling decision-making for subsequent steps based on the recorded inter-agent communication. LangChain \cite{githubGitHubLangchainailangchain} enhances short-term memory efficiency by encapsulating crucial information from each interaction and preserving the most recurrent interactions.

\paragraph{Long-term Memory} Long-term memory stores and regulates substantial volumes of knowledge, experiential data, and historical records. An agent utilizing long-term memory may incorporate interaction with external knowledge bases, databases, or other information sources. The design of external memory can leverage techniques such as knowledge graphs \cite{wang2017knowledge}, vector databases \cite{singla2021experimental}, relational database queries, or API calls to engage with external data sources. Voyager \cite{wang2023voyager} employs a perpetually expanding skill repository for storing and retrieving complex behaviors. In GITM \cite{zhu2023ghost}, memory primarily aids in extracting the most pertinent textual knowledge from an external knowledge base, which the long-term memory subsequently utilizes to identify necessary materials, tools, and related information. To augment agent performance, the ExpeL \cite{zhao2023expel} agent preserves experiences across multiple tasks. In Reflexion \cite{shinn2023reflexion}, experiences acquired through self-reflection are conserved in long-term memory and influence future actions. MemGPT \cite{packer2023memgpt} is an intelligent system adept at managing diverse memory hierarchies, effectively providing an extended context within the limited context window of LLMs and utilizing interrupts to manage control flow between itself and the user.

Short-term memory can encapsulate and generalize vital information, which is then dynamically stored in long-term memory. As demonstrated by Generative Agents \cite{park2023generative}, the agent sustains its internal state by archiving and updating its experiences, generating natural language by aligning its experiences with the language representations of LLMs, and consistently amassing new experiences and integrating them with existing ones. The agent's memory undergoes evolution over time and can be dynamically accessed to represent the agent's current state.

\paragraph{Memory Retrieval} 
Retrieval-augmented generation~\cite{lewis2020retrieval} can combine an information retrieval component with an LLM and produce a more reliable output. The retrieval objectives can be represented with a memory, i.e., a knowledge library. Memory retrieval is paramount for the proficient access and administration of memories. In the context of LLM-based agents, memory retrieval can be facilitated through online learning and adaptive modification. When formulating memory retrieval methods, techniques such as online reinforcement learning, multitask learning, or attention mechanisms can enable real-time updates and adjustments to model parameters. LaGR-SEQ \cite{karimpanal2023lagr} introduces SEQ (Sample Efficient Query), which trains a secondary RL-based agent to determine when to query the LLM for solutions. REMEMBER \cite{zhang2023large} equips LLMs with long-term memory, empowering them to draw from past experiences, and introduces Reinforcement Learning and Experience Memory to update memories. Synapse \cite{zheng2023synapse} purges task-irrelevant information from the raw state, enabling more samples within a restricted context. It generalizes to novel tasks by storing sample embeddings and retrieving them via similarity search. \citet{kang2023think} discusses the characteristics of distributed memory storage in the human brain. It proposes the construction of an internal memory module, DT-Mem, which allows agents to store and retrieve information relevant to various downstream tasks. \citet{wang2023jarvis1} utilizes a multimodel memory to store the agent's collected interaction experiences and use the embodied RAG to make the agents self-improve by exploring the open-world Minecraft.

\vspace{.5em}

Utilizing the approaches above, it is feasible to devise memory types and retrieval techniques tailored for LLM-based agents. It is imperative to highlight that LLM-based agents can encompass both memory categories concurrently. The judicious selection of pertinent memory classification and retrieval mechanisms can bolster LLM-based agents in proficiently storing, administering, and expeditiously extracting data while addressing challenges and accomplishing tasks, thus augmenting their efficacy and adaptability.

\subsubsection{Rethinking} The capacity for introspection in an LLM-based agent, denoted as its rethinking ability, encompasses evaluating prior decisions and subsequent environmental feedback. This faculty permits an LLM-based agent to thoroughly examine its behavior, decision-making, and learning processes, augmenting its intelligence and adaptability.

Contemporary investigations on LLM-based agent rethinking can be extensively classified according to learning methodologies, which include In-Context Learning, Supervised Learning, Reinforcement Learning, and Modular Coordination approaches.

\begin{figure*}[!ht]
\scriptsize
\begin{adjustbox}{width=\textwidth}
\begin{forest}
  for tree={
    forked edges,
    grow'=0,
    draw,
    rounded corners,
    node options={align=center},
    text width=3cm,
    s sep=5pt,
    calign=edge midpoint, 
    fill=white,
    drop shadow,
  }
  [Rethinking Capability of LLM-based Agents, fill=gray!45
[In-Context Learning Methods
[ReAct\cite{yao2022react}]
[Reflexion\cite{shinn2023reflexion}]
]
[Supervised Learning Methods
[CoH\cite{liu2023chain}]
[Process Supervision\cite{lightman2023let}]
[Introspective Tips\cite{chen2023introspective}]
[Hinting Approach\cite{zhou2023solving}]
]
[Reinforcement Learning Methods
[Retroformer\cite{yao2023retroformer}]
[REMEMBER\cite{zhang2023large}]
[Dialogue Shaping\cite{zhou2023dialogue}]
[REX\cite{murthy2023rex}]
[ICPI\cite{brooks2023large}]
]
[Modular Coordination Methods
[DIVERSITY\cite{li2022advance}]
[DEPS\cite{wang2023describe}]
[PET\cite{wu2023plan}]
[Three-Part System\cite{dasgupta2023collaborating}]
]
]
\end{forest}
\end{adjustbox} 
\caption{Typology of the Rethinking capability.}
\label{fig:rethink_mindmap}
\end{figure*}
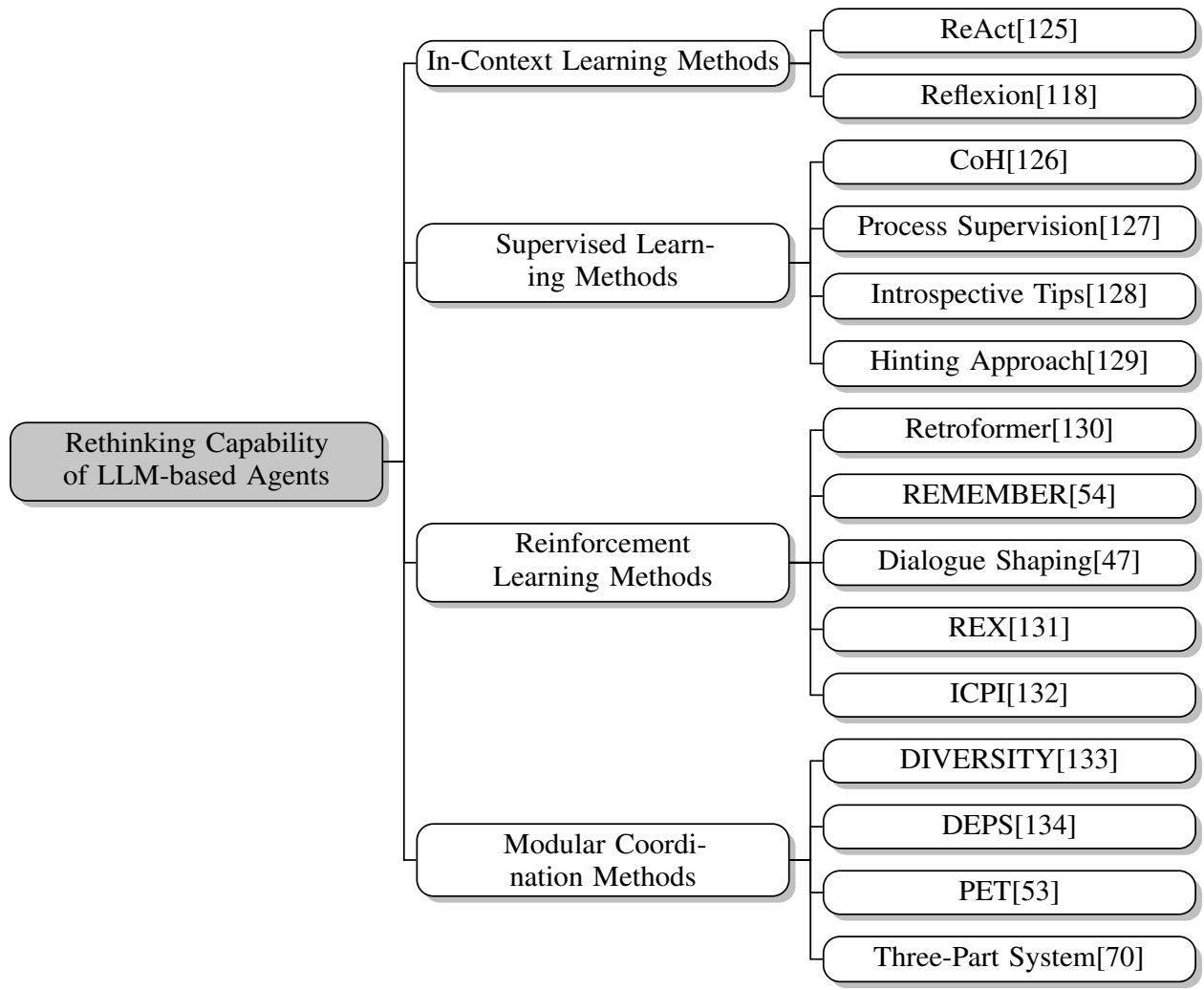

\paragraph{In-Context Learning Methods} As delineated in Section \ref{framework_single_agent_planning}, In-Context Learning (ICL) leverages task-specific linguistic prompts and instances for reinforcement. ReAct \cite{yao2022react} implements an interactive paradigm, alternating between generating task-related linguistic reasoning and actions, thereby fostering a synergistic enhancement of the language model's reasoning and action proficiencies. This approach exhibits generality and adaptability in addressing tasks requiring diverse action spaces and reasoning. Reflexion \cite{shinn2023reflexion} computes heuristics after each action and ascertains whether to reset the environment based on self-reflection, consequently bolstering the agent's reasoning capabilities.

\paragraph{Supervised Learning Methods} Supervised learning typically hinges on diverse sources, encompassing LLMs, human expertise, code compilers, and external knowledge. CoH \cite{liu2023chain} exploits a sequence of prior outputs annotated with feedback to foster model self-enhancement. This technique employs supervised fine-tuning, positive and negative grading, and experience replay to augment performance. \citet{lightman2023let} experimentally substantiates that process supervision surpasses outcome supervision in mathematical reasoning tasks, and active learning considerably boosts the efficacy of process supervision. Introspective Tips \cite{chen2023introspective} introduces a self-examination framework predicated on past trajectories or expert demonstrations, generating succinct yet valuable insights for strategy optimization. \citet{zhou2023solving} advocates a hinting methodology founded on explicit code-based self-verification to refine the mathematical reasoning prowess of the GPT-4 code interpreter. Moreover, it incorporates a Diversity Verifier on the Reasoning Step to strengthen the agent's reasoning aptitudes further.

\paragraph{Reinforcement Learning Methods} Reinforcement learning emphasizes the enhancement of parameters by acquiring knowledge from historical experiences. Retroformer \cite{yao2023retroformer} ameliorates agents by learning from retrospective models and employing policy gradients to modulate the LLM-based agent's prompts autonomously. REMEMBER \cite{zhang2023large} introduces a novel semi-parametric reinforcement learning methodology that amalgamates reinforcement learning and experience memory to update memory and augment capabilities via experiential analogies. \citet{zhou2023dialogue} offers a framework for shaping dialogues, expediting agent convergence to optimal strategies by extracting pertinent information from non-player characters (NPCs) and transmuting it into a knowledge graph. REX \cite{murthy2023rex} incorporates an auxiliary reward layer and assimilates concepts akin to Upper Confidence Bound scores, culminating in more robust and efficient AI agent performance. ICPI \cite{brooks2023large} demonstrates the capacity to execute RL tasks without expert demonstrations or gradients by iteratively updating prompt content through trial-and-error interactions within the RL environment. \citet{liu2023reason} integrates agent planning and actions by utilizing learning and planning in Bayesian adaptive Markov decision processes (MDP). In this approach, LLM constructs updated posteriors of unknown environments from memory buffers for learning while generating optimal trajectories that maximize value functions for multiple future steps in planning. \citet{wang2023adapting} proposes a technique that enables LLM-based agents to undergo perpetual improvement through iterative exploration and Proximal Policy Optimization (PPO) \cite{schulman2017proximal} training in interaction with the environment and other agents. This method additionally facilitates the integration of short-term experiences into long-term memory.

\paragraph{Modular Coordination Methods} Modular coordination methods typically encompass multiple modules operating in concert to facilitate planning and introspection for LLM-based agents. DIVERSITY \cite{li2022advance} investigates various prompts to augment the diversity of reasoning pathways. By incorporating a verifier to discern between favorable and unfavorable responses, it attains enhanced weighted voting and employs diversity verification to ascertain the correctness of each step. The DEPS \cite{wang2023describe} framework interacts with LLM planners through descriptors, interpreters, and goal selectors, improving overall success rates. PET \cite{wu2023plan} leverages LLM knowledge to streamline control problems for embodied agents. This framework includes planning, elimination, and tracking modules to accomplish higher-level subtasks. \citet{dasgupta2023collaborating} examines the integration of planners, actors, and reporters into a tripartite system. This system demonstrates generalization capabilities in distributed learning, investigates failure scenarios, and delineates how reinforcement learning trains individual components to enhance performance.

\vspace{.5em}
These methods and frameworks optimize the performance of LLM-based agents through environmental feedback, self-learning, and reflection. They have achieved significant advancements in enhancing the capabilities of LLM-based agents in reflection and re-planning.

\subsubsection{Environments} The LLM-based agents can interact and learn from various environments through environmental feedback. These environments can broadly be computer, gaming, code, real-world, and simulation environments.

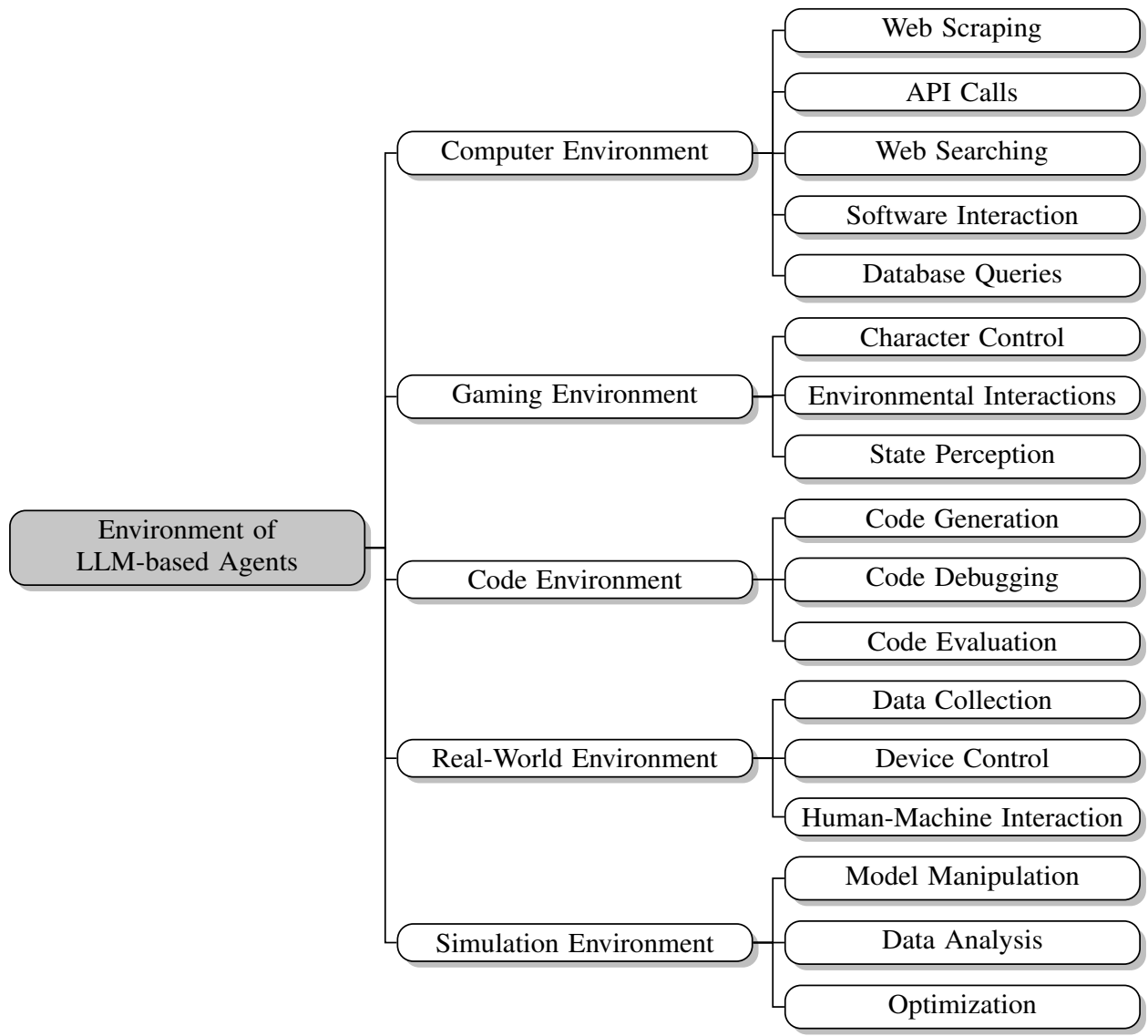
\begin{figure*}[!ht]
\scriptsize
\begin{adjustbox}{width=\textwidth}
\begin{forest}
  for tree={
    forked edges,
    grow'=0,
    draw,
    rounded corners,
    node options={align=center},
    text width=3cm,
    s sep=5pt,
    calign=edge midpoint, 
    fill=white,
    drop shadow,
  }
  [Environment of LLM-based Agents, fill=gray!45
[Computer Environment
[Web Scraping]
[API Calls]
[Web Searching]
[Software Interaction]
[Database Queries]
]
[Gaming Environment
[Character Control]
[Environmental Interactions]
[State Perception]
]
[Code Environment
[Code Generation]
[Code Debugging]
[Code Evaluation]
]
[Real-World Environment
[Data Collection]
[Device Control]
[Human-Machine Interaction]
]
[Simulation Environment
[Model Manipulation]
[Data Analysis]
[Optimization]
]
]
\end{forest}
\end{adjustbox} 
\caption{Typology of the Environment.}
\label{fig:environment_mindmap}
\end{figure*}

\paragraph{Computer Environment}
LLM-based agents engage with websites, APIs, databases, and applications across computer, web, and mobile contexts in computational environments. The modes of interaction encompass:
\begin{itemize}
    \item \textbf{Web Scraping}: Gleaning information from websites to acquire essential data and knowledge.
    \item \textbf{API Calls}: Employing Web APIs to access or transmit data, fostering interactions with online services.
    \item \textbf{Web Searching}: Utilizing search engines to discover pertinent information and resources for problem-solving or task completion.
    \item \textbf{Software Interaction}: Manipulating and interfacing with software applications, ranging from word processors to graphic design tools, to efficiently execute tasks.
    \item \textbf{Database Queries}: Access and update databases directly, enabling real-time data handling and processing.
\end{itemize}

Contemporary research introduces methodologies such as RCI \cite{kim2023language}, which guides language models to perform computational tasks via natural language commands. WebArena \cite{zhou2023webarena} offers an independent, self-hosted web environment for constructing autonomous agents. WebGPT \cite{nakano2021webgpt} capitalizes on search engines for document retrieval, enabling end-to-end imitation and reinforcement learning to optimize retrieval and aggregation while generating responses that reference web-retrieved information. Mobile-Env \cite{zhang2023mobileenv} permits agents to observe screenshots and view frameworks of the Android operating system, enabling actions such as tapping the screen or inputting commands to interact with Android applications. SheetCopilot \cite{li2023sheetcopilot} facilitates interaction with spreadsheets using natural language.

\paragraph{Gaming Environment}
LLM-based agents interact with virtual characters, objects, and settings in gaming environments. Interaction methodologies in gaming contexts include:
\begin{itemize}
    \item \textbf{Character Control}: Directing in-game characters by issuing commands (e.g., move, jump, attack).
    \item \textbf{Environmental Interactions}: Engaging with objects in the game environment (e.g., pick up, use, place) to complete tasks.
    \item \textbf{State Perception}: Gathering status information from the game environment (e.g., character position, item count) for decision-making and planning.
\end{itemize}

Prominent applications encompass DECKARD \cite{nottingham2023embodied}, which deploys LLM-guided exploration for devising tasks within the Minecraft game. VOYAGER \cite{wang2023voyager} constitutes a Minecraft-based LLM-driven lifelong learning agent that persistently explores the world, acquires various skills, and uncovers discoveries. GITM \cite{zhu2023ghost} employs an "indirect mapping" approach to translate long-term and intricate goals into a series of low-level keyboard and mouse actions, facilitating efficient and adaptable operations in the Minecraft game. AgentSims \cite{lin2023agentsims} generates a virtual town with diverse buildings and residents, streamlining task design and addressing challenges researchers might encounter due to varying backgrounds and programming expertise levels. LLM-Deliberation \cite{abdelnabi2023llmdeliberation} establishes a versatile testing platform for text-based, multi-agent, multi-issue, and semantically rich negotiation games. Moreover, the platform enables effortless adjustment of difficulty levels.

\paragraph{Coding Environment}

The coding environment enables LLM-based agents to compose, modify, and execute code for various tasks, from coding to verifying reasoning through code. Interaction methodologies in the code environment encompass code generation, code debugging, and code evaluation. Code generation produces code snippets or complete programs based on task requirements. Code debugging identifies and rectifies errors or issues in the code. Code evaluation executes code and assesses its performance, optimizing and refining it based on runtime error messages or output.

LLift \cite{li2023hitchhikers} constitutes a fully automated agent that interfaces with static analysis tools and LLM to address challenges such as error-specific modeling, extensive problem scopes, and LLM's non-determinism. MetaGPT \cite{hong2023metagpt} incorporates human workflows into LLM-driven collaboration, employing Standard Operating Procedures (SOPs) as prompts to facilitate structured coordination. Similarly, \citet{dong2023self} introduces a self-collaboration framework involving multiple LLM roles for auto-generating code. Within this framework, distinct roles assume analyst, programmer, tester, etc., forming a collaborative team to achieve code generation tasks. ChatDev \cite{qian2023communicative} represents a virtual chat-driven software development company that segments the development process into four discrete sequential stages based on the waterfall model: design, coding, testing, and documentation. CSV \cite{zhou2023solving} augments mathematical reasoning abilities by prompting the interpreter to utilize code for self-verification, enhancing solution confidence by indicating verified states.

\paragraph{Real-World Environment}

LLM-based agents can interact with real-world devices, sensors, and actuators, facilitating their operation in real-world scenarios. The interaction methodologies in these situations include:
\begin{itemize}
\item \textbf{Data Collection}: LLM-based agents can accumulate real-time data from sensors such as cameras and microphones, which are then employed for analysis and decision-making.
\item  \textbf{Device Control}: Regarding device control, LLM-based agents can manipulate actuators like robotic arms and drones by transmitting control signals, thereby accomplishing specific tasks.
\item \textbf{Human-Machine Interaction}: Regarding human-machine interaction, LLM-based agents excel in engaging in natural language communication with human users, enabling the reception of instructions, provision of feedback, and response to queries.
\end{itemize}

\citet{di2023towards} introduces a language-centric reasoning toolkit framework tested in a sparse reward robot manipulation environment where robots execute tasks like stacking objects. TaPA \cite{wu2023embodied} presents an embedded task-planning agent for real-world planning under physical scene constraints. The SimBot challenge in the Alexa Prize project \cite{shi2023alexa} aims to construct robot assistants capable of completing tasks in a simulated physical environment. \citet{zheng2023synergizing} proposes 23 heuristic methods for guiding LLM-based agents in collaborating and co-creating services alongside humans.

\paragraph{Simulation Environment}

LLM-based agents use virtual models representing real-world systems or processes in simulation environments, such as economic markets, physical environments, and transportation systems. Interaction methodologies in simulation environments include:

\begin{itemize}
    \item \textbf{Model Manipulation}: Adjusting parameters or variables within the simulation model to explore various scenarios and analyze their outcomes.
    \item \textbf{Data Analysis}: Collect and analyze data generated by the simulation to identify patterns, trends, and insights that can inform decision-making.
    \item \textbf{Optimization}: Applying optimization algorithms to determine the best course of action within the simulated environment, considering constraints and objectives.
\end{itemize}

In recent research, TrafficGPT \cite{zhang2023trafficgpt} demonstrates the capability to perform traffic flow analysis and address questions within the traffic simulation environment SUMO \cite{lopez2018microscopic}. \citet{li2023you} examines the behavioral characteristics of social agents in simulated social platforms. \citet{horton2023large} investigates the behavior of LLM-based agents in economic simulation scenarios and compares the differences between agent and actual human behavior. AucArena \cite{chen2023money} is a simulation environment for auctions wherein agents must consider resource and risk management factors.

These simulation environments provide a controlled yet realistic context for LLM-based agents to learn, experiment, and develop solutions applicable to real-world scenarios, facilitating the transfer of knowledge and skills from the virtual domain to real-life applications.

\vspace{.5em}
In summary, the LLM-based agent learns and applies knowledge through natural language interaction and environmental feedback across various environments, offering robust solutions for different tasks.

\subsubsection{Action}

The action capabilities of an LLM-based agent pertain to the performance of actions or the employment of tools. Such agents' predominant mode of interaction is typically through text generation, facilitating communication with the external environment, a characteristic reminiscent of the Generative Agents \cite{park2023generative}. An alternative methodology incorporates the LLM or the agent-employing tools, encompassing APIs, calculators, code interpreters, or actions within a physical environment through text-based directives. This further extends to the strategic planning and deployment of tools, which may necessitate the development of new tools for their implementation.

\begin{figure*}[!ht]
\scriptsize
\begin{adjustbox}{width=\textwidth}
\begin{forest}
  for tree={
    forked edges,
    grow'=0,
    draw,
    rounded corners,
    node options={align=center},
    text width=3cm,
    s sep=5pt,
    calign=edge midpoint, 
    fill=white,
    drop shadow,
  }
  [Actions of LLM-based Agents, fill=gray!45
[Tool Employment
[MRKL\cite{karpas2022mrkl}]
[TALM\cite{parisi2022talm}]
[ToolFormer\cite{schick2023toolformer}]
[HuggingGPT\cite{shen2023hugginggpt}]
[Chameleon \cite{lu2023chameleon}]
[Gorilla \cite{patil2023gorilla}]
[RestGPT\cite{song2023restgpt}]
[D-Bot\cite{zhou2023llm}]
[Chameleon\cite{lu2023chameleon}]
[AVIS\cite{hu2023avis}]
]
[Tool Planning
[ChatCoT\cite{chen2023chatcot}]
[TPTU\cite{ruan2023tptu}]
[ToolLLM\cite{qin2023toolllm}]
[Gentopia\cite{xu2023gentopia}]
]
[Tool Creation
[\citet{cai2023large}]
[CRAFT\cite{yuan2023craft}]
]
]
\end{forest}
\end{adjustbox} 
\caption{Typology of the Actions.}
\label{fig:action_mindmap}
\end{figure*}
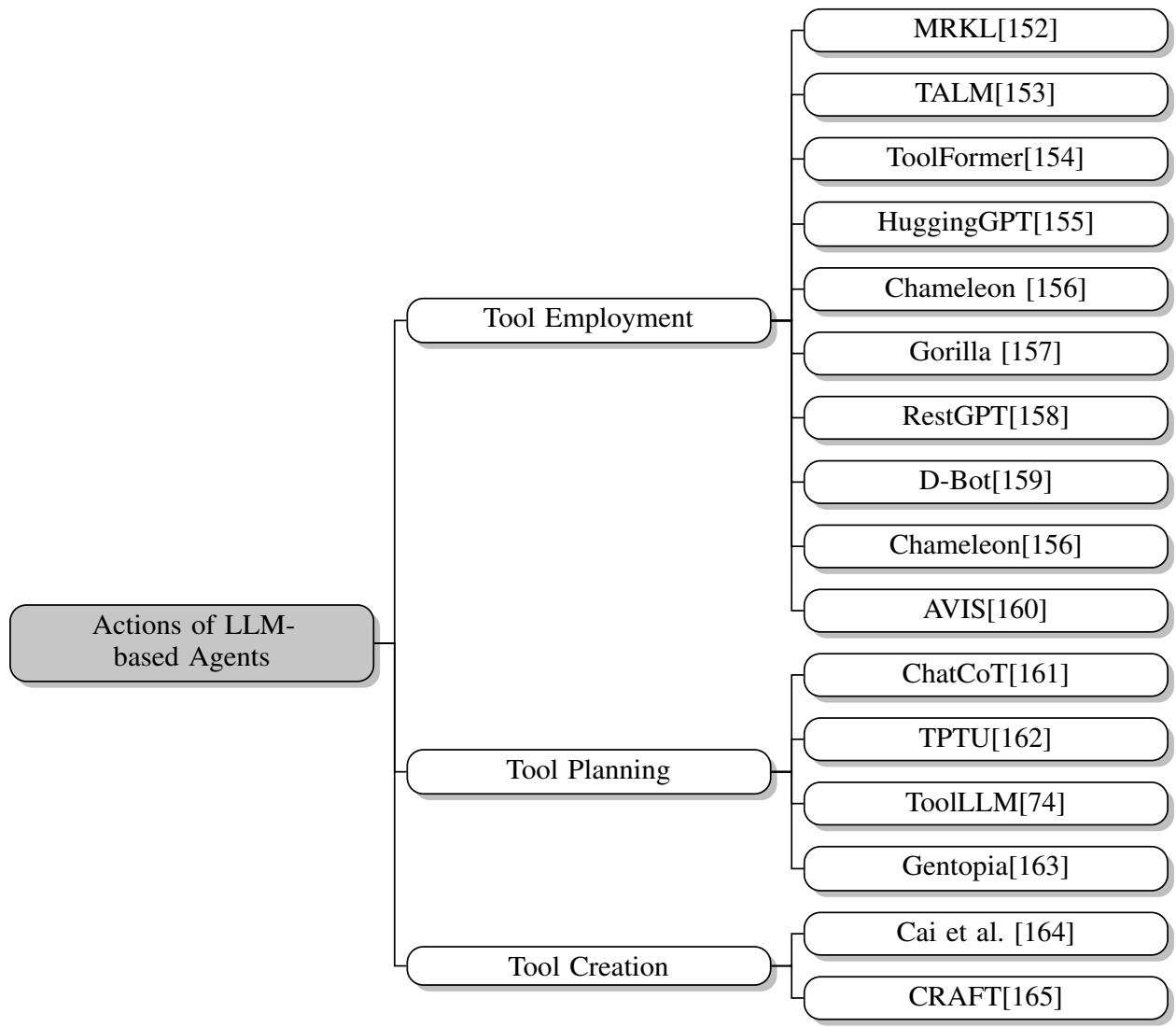

\paragraph{Tool Employment}

MRKL \cite{karpas2022mrkl} integrates LLM and external tools to address intricate problems. This encompasses the construction of modules and routers and the routing of natural language queries. TALM \cite{parisi2022talm} connects language models with tools, facilitating text-to-text API connections. ToolFormer \cite{schick2023toolformer} exemplifies the LLM's capacity to leverage external tools, augmenting performance across various tasks. HuggingGPT \cite{shen2023hugginggpt} combines multiple AI models and tools for task planning and execution, including text classification and object detection. Tool Learning with Foundation Models explores tool learning, presenting a generalized framework that merges foundational models and toolsets to achieve efficient task execution. Gorilla \cite{patil2023gorilla} delves into LLM's applications in API calls and program synthesis, context learning, and task decomposition to enhance performance. RestGPT \cite{song2023restgpt} is a method that connects LLM with RESTful APIs to address user requests, including online planning and API execution. TaskMatrix.AI \cite{liang2023taskmatrixai} can comprehend inputs in text, images, videos, audio, and code and subsequently generates code that invokes APIs to accomplish tasks. D-Bot \cite{zhou2023llm} offers database maintenance suggestions covering knowledge detection, root cause analysis, and multi-LLM collaboration. 

Chameleon \cite{lu2023chameleon} employs various tools to address challenges and utilizes a natural language planner to select and combine modules stored in the inventory, thereby constructing solutions. AVIS \cite{hu2023avis} is an autonomous visual information-seeking system that leverages LLMs to dynamically formulate strategies for utilizing external tools and examining their output results, thus acquiring essential knowledge to provide answers to the posed questions.

\paragraph{Tool Planning}

ChatCoT \cite{chen2023chatcot} models chain-like thinking into multi-turn dialogues, improving complex task handling through tool-aided reasoning. TPTU \cite{ruan2023tptu} introduces a task execution framework comprising task instructions, design prompts, toolkits, LLM, results, and task planning and tool utilization capability. ToolLLM \cite{qin2023toolllm} develops a Decision Tree based on Depth-First Search, enabling LLMs to evaluate multiple API-based reasoning paths and expand the search space. Gentopia \cite{xu2023gentopia} is a framework allowing flexible customization of agents through simple configuration, seamlessly integrating various language models, task formats, prompt modules, and plugins into a unified paradigm.

\paragraph{Tool Creation}

\citet{cai2023large} presents a tool creation and utilization framework, generating tools suitable for diverse tasks. This covers staged tool generation and task execution. CRAFT \cite{yuan2023craft} is a framework designed for developing and retrieving general-purpose tools, enabling the generation of specialized toolkits tailored for specific tasks. LLM can extract tools from these toolkits to address complex tasks.

\subsection{LLM-based Multi-Agent System}\label{framework-multi-agent}

\subsubsection{Relationship of  Multi-Agent Systems} 

In LLM-based Multi-Agent Systems (MAS), many agents engage in collaboration, competition, or hierarchical organization to execute intricate tasks. These tasks could range from search and optimization, decision support, and resource allocation to collaborative generation or control. The interrelationships between agents in these systems are of paramount importance as they govern the mechanisms of interaction and cooperation among agents. Similarly, these inter-agent relationships can be extrapolated to LLM-based MAS. Currently, most research in LLM-based MAS primarily focuses on the cooperative and competitive dynamics between agents.

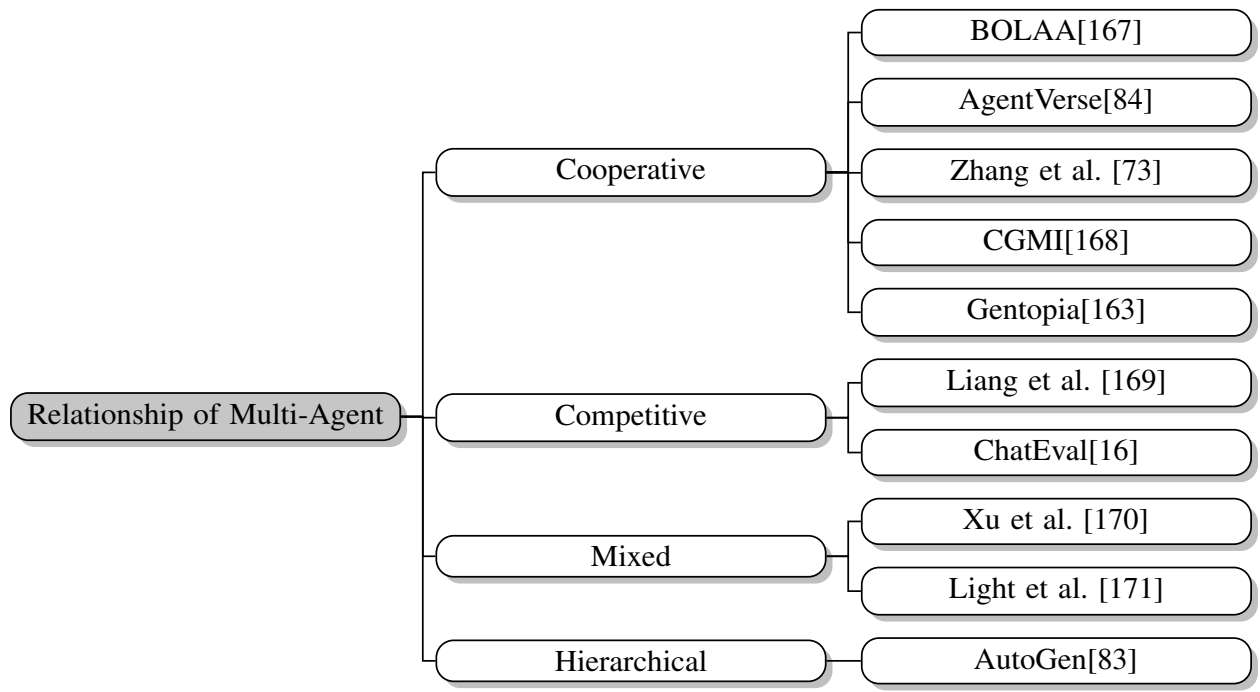
\begin{figure*}[!ht]
\scriptsize
\begin{adjustbox}{width=\textwidth}
\begin{forest}
  for tree={
    forked edges,
    grow'=0,
    draw,
    rounded corners,
    node options={align=center},
    text width=3cm,
    s sep=5pt,
    calign=edge midpoint, 
    fill=white,
    drop shadow,
  }
  [Relationship of Multi-Agent, fill=gray!45
[Cooperative
[BOLAA\cite{liu2023bolaa}]
[AgentVerse\cite{chen2023agentverse}]
[\citet{zhang2023building}]
[CGMI\cite{jinxin2023cgmi}]
[Gentopia\cite{xu2023gentopia}]
]
[Competitive
[ \citet{liang2023encouraging}]
[ChatEval\cite{chan2023chateval}]
]
[Mixed
[\citet{xu2023exploring}]
[ \citet{light2023text}]
]
[Hierarchical
[AutoGen\cite{wu2023autogen}]
]
]
\end{forest}
\end{adjustbox} 
\caption{Relationship of Multi-Agent System.}
\label{fig:coordination_mindmap}
\end{figure*}

\paragraph{Cooperative Relationship}
In cooperative relationships, scholarly attention is predominantly directed toward role and task allocation strategies and algorithms for collaborative decision-making. Such methodologies can enhance the efficacy of agent collaboration, resulting in improved overall system performance. SPP \cite{wang2023unleashing} facilitates multi-turn dialogue via multi-persona self-cooperation, transforming a singular LLM into a cognitive synergist. Generative Agents \cite{park2023generative} employ LLM-based agents to emulate plausible human behavior, thus fostering cooperation amongst agents. CAMEL \cite{li2023camel} realizes multi-turn dialogue cooperation between AI assistants and AI users through task-oriented role-playing. MetaGPT \cite{hong2023metagpt} integrates effective workflows into LLM-driven multi-agent collaboration programming methodologies, enabling collaboration amongst diverse roles. ChatDev \cite{qian2023communicative} utilizes multiple LLM-based agents for dialogue task resolution, expediting LLM application development.

Drawing inspiration from Minsky's society of mind~\cite{minsky1988society}, NLSOM~\cite{zhuge2023mindstorms} introduces the notion of natural language-based societies of mind (NLSOMs), comprising multiple LLMs and other neural network-based experts that communicate through a natural language interface. This approach is applied to tackle complex tasks across various scenarios. \citet{zou2023wireless} enables collaboration among device-side LLMs. Regarding embodied MAS, RoCo \cite{mandi2023roco} employs LLMs for high-level communication and low-level path planning, facilitating multi-robot collaboration. InterAct \cite{chen2023interact} assigns roles such as inspectors and classifiers, achieving notable success rates in AlfWorld \cite{shridhar2020alfworld}. AutoAgents \cite{chen2023auto} can adaptively generate and coordinate multiple specialized agents, forming an AI team to accomplish goals based on various tasks.

In the study of LLM-based MAS frameworks, BOLAA \cite{liu2023bolaa} devises an architecture for orchestrating multi-agent strategies, augmenting the action interaction capabilities of individual agents. AgentVerse \cite{chen2023agentverse} offers a versatile framework that simplifies creating custom multi-agent environments for LLMs. \citet{zhang2023building} proposes a novel framework that empowers embodied agents to plan, communicate, and collaborate, efficiently completing long-term tasks with other embodied agents or humans. CGMI \cite{jinxin2023cgmi} is a configurable generic multi-agent interaction framework that harnesses LLM capabilities to address agent performance challenges in specific tasks and simulate human behavior. Gentopia \cite{xu2023gentopia} is a framework that allows for flexible customization of agents through simple configuration, seamlessly integrating various language models, task formats, prompt modules, and plugins into a unified paradigm. DyLAN \cite{liu2023dynamic} introduces a task-based dynamic framework that enables multiple agents to interact and presents an automatic agent team optimization algorithm based on unsupervised metrics. This algorithm selects the most effective agents according to the contributions of each agent.

\paragraph{Competitive Relationship} 
In competitive relationships, considerations encompass designing effective competitive strategies, information concealment techniques, and adversarial behavior. These techniques can assist agents in gaining an advantage in competition, thereby achieving their goals. \citet{liang2023encouraging} enhances task-solving capabilities through a multi-agent debate framework. ChatEval \cite{chan2023chateval} employs a multi-agent approach to facilitate a group of LLMs collaborating with various intelligent opponents, leveraging their respective abilities and expertise to improve the efficiency and effectiveness of processing complex tasks.

\paragraph{Mixed Relationship}
Agents must balance cooperation and competition in mixed relationships to achieve their goals. Currently, research on mixed relationships in LLM-based MAS focuses on the design of collaborative competition algorithms, which is a crucial topic. These techniques can aid agents in making effective decisions in complex environments. \citet{xu2023exploring} enables multiple LLM-based agents to participate in the Werewolf game, with each agent cooperating or betraying other agents to fulfill their role's goals under asymmetric information conditions. Similarly, \citet{light2023text} enables LLM-based agents to participate in the Avalon game, where each agent must make decisions during dynamically evolving game stages and engage in negotiations involving cooperation or deception with other agents to accomplish the objectives of their assigned roles. Drawing inspiration from human behavior, Corex \cite{sun2023corex} incorporates various collaboration paradigms, such as debate, review, and retrieval modes. Collectively, these modes strive to augment the reasoning process's authenticity, fidelity, and reliability.

\paragraph{Hierarchical Relationship}
Researchers focus on developing efficient hierarchical control structures, information transmission mechanisms, and task decomposition methods in hierarchical relationships. These techniques enable agents to collaborate effectively across different levels, augmenting the system's overall performance. Hierarchical relationships typically manifest as a tree structure, wherein parent-node agents undertake the task decomposition process and assign tasks to child-node agents. The latter agents adhere to the arrangements set by their corresponding parent nodes and provide summarized information in return. AutoGen \cite{wu2023autogen} employs diverse agents to tackle tasks, such as code generation and text writing, utilizing task decomposition through dialogue. Presently, research on hierarchical relationships in LLM-based MAS is still developing, with only a limited number of levels being explored.

\vspace{.5em} 

In forthcoming research endeavors, the utilization of game theory, auction mechanisms, and negotiation techniques holds promise in tackling the challenges associated with task allocation problems among cooperative agents. Furthermore, Distributed Constraint Optimization Problems (DCOP) present a substantial framework for investigating collaborative decision-making within cooperative agents. In the context of other relationship types, cooperative games and Multi-Objective Reinforcement Learning (MORL) emerge as pivotal frameworks for exploring the delicate equilibrium between cooperation and competition. These established research frameworks can also be adapted and refined within LLM-based MAS.

\subsubsection{Planning Type}

In the domain of MAS, planning constitutes a crucial component as it enables the orchestration of multiple agents in pursuit of shared goals. Numerous planning methodologies have been put forth, each exhibiting unique merits and constraints. Analogous to the notion of Centralized Training Decentralized Execution (CTDE) \cite{lowe2017multi} in Multi-Agent Reinforcement Learning, this investigation delves into two primary planning paradigms: Centralized Planning Decentralized Execution (CPDE) and Decentralized Planning Decentralized Execution (DPDE).

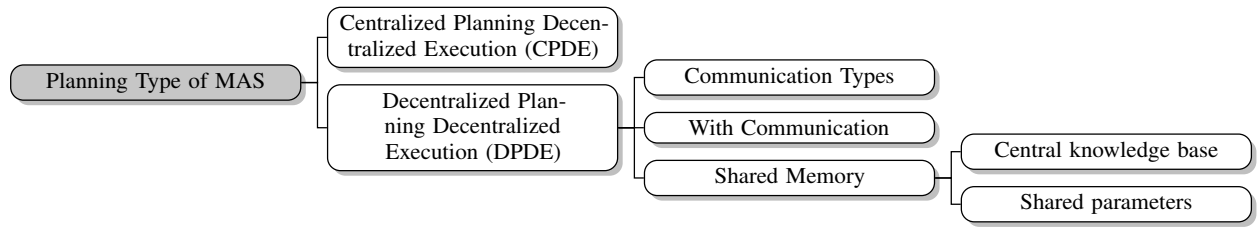
\begin{figure*}[!ht]
\scriptsize
\begin{adjustbox}{width=\textwidth}
\begin{forest}
  for tree={
    forked edges,
    grow'=0,
    draw,
    rounded corners,
    node options={align=center},
    text width=3cm,
    s sep=5pt,
    calign=edge midpoint, 
    fill=white,
    drop shadow,
  }
  [Planning Type of MAS, fill=gray!45
[Centralized Planning Decentralized Execution (CPDE)
]
[Decentralized Planning Decentralized Execution (DPDE)
[Communication Types
]
[With Communication
]
[Shared Memory
[Central knowledge base]
[Shared parameters]
]
]
]
]
\end{forest}
\end{adjustbox} 
\caption{Typology of the MAS Planning Type.}
\label{fig:planningtype_mindmap}
\end{figure*}

\paragraph{Centralized Planning Decentralized Execution (CPDE)}\label{CPDE}

In the CPDE paradigm, a centralized LLM is responsible for planning on behalf of all agents encompassed within the system. This requires the LLM to consider all agents' objectives, capabilities, and constraints to devise appropriate action plans for them. As highlighted in \citet{gong2023mindagent}, the planner must concurrently manage multiple agents, circumvent potential conflicts, and coordinate them to accomplish a shared goal necessitating intricate collaborations. Upon finalizing the planning, each agent independently carries out its designated tasks without further interaction with the central LLM. The merit of this approach resides in optimizing overall performance at a global scale, as the central LLM can consider the needs and resources of all agents. \citet{li2023semantically} develops SAMA in both overcooked \cite{carroll2019utility} and MiniRTS \cite{hu2019hierarchical} multi-agent environments by employing a centralized LLM to facilitate goal generation, goal decomposition, goal allocation, and replanning with self-reflection.

However, CPDE also exhibits certain limitations. Firstly, the centralized planning process may lead to escalated computational complexity, particularly when managing numerous agents and complex tasks. Secondly, given that all agents depend on a single LLM for planning, the system may be susceptible to single-point failures and communication delays. Lastly, CPDE might not be well-suited for situations demanding real-time responsiveness and heightened adaptability, as the central LLM may be incapable of swiftly responding to environmental changes.

\paragraph{Decentralized Planning Decentralized Execution (DPDE)}\label{DPDE}

Contrasting with CPDE, DPDE systems incorporate individual LLMs responsible for action planning in each agent. As a result, every agent can independently formulate plans based on its objectives, capabilities, and local information. During the execution phase, agents can enhance collaboration by coordinating their actions through local communication and negotiation.

The benefits of DPDE encompass increased robustness and scalability as each agent independently plans and executes, thereby alleviating the computational burden on the central LLM. Furthermore, DPDE systems typically demonstrate greater adaptability, as each agent can promptly modify its behavior according to local information. This attribute renders DPDE systems more appropriate for dynamic and uncertain environments.

However, the constraints of DPDE include potential challenges in attaining global optimality, as each agent's planning is contingent on local information. Moreover, coordination and communication overhead may become considerable in large-scale systems, potentially influencing overall performance.

Information exchange between agents is vital for promoting cooperation and collaboration in such systems. The ensuing discussion delineates three categories of information exchange between agents in DPDE systems:

\subparagraph{\quad Information Exchange Without Communication}

In this modality, agents abstain from direct communication. Each agent independently plans and executes, depending solely on local information and observations to accomplish tasks. The merit of this approach lies in minimal communication overhead, as agents are not required to exchange information. Moreover, this method might be the sole viable option in environments characterized by limited or unreliable communication.

Nonetheless, the lack of communication may give rise to suboptimal collaboration among agents, as they cannot share information, coordinate actions, or resolve conflicts. In certain instances, this may culminate in inefficient behavior and a deterioration in overall performance.

\subparagraph{\quad Information Exchange With Communication}

In this modality, agents engage in information exchange and action coordination through explicit communication. Communication can assume diverse forms, including message passing, broadcasting, or point-to-point communication. Agents can disseminate observations, goals, plans, and other relevant information by communicating bolstering collaboration and overall performance.

However, communication may incur supplementary overhead, encompassing communication delays, bandwidth utilization, and receiving information processing. Furthermore, in environments characterized by unreliable or limited communication, this approach may confront obstacles, such as lost messages or delayed updates.

\subparagraph{\quad Information Exchange With Shared Memory}

In this modality, agents exchange information via shared memory, a centralized data structure accessible and modifiable by all agents within the system. Agents accomplish information sharing and collaboration by storing and retrieving information in shared memory.

Shared memory presents several merits, such as streamlining communication, as agents need not directly transmit and receive messages. Furthermore, it offers a unified information representation and access mechanism, simplifying the system's design and implementation.

However, shared memory exhibits certain limitations. Firstly, contention and synchronization issues may emerge as all agents necessitate access to and modification of shared memory. Secondly, shared memory could impede the system's scalability, requiring consistency among all agents. Lastly, implementing shared memory in distributed and mobile agent environments may confront technical challenges, such as ensuring data consistency and managing concurrency control.

In contemporary research, two forms of shared memory can be identified:

\begin{itemize}
    \item \textbf{Central Knowledge Base}: A central knowledge base can be established to store and manage the shared knowledge of each agent. This knowledge base can be a database, a knowledge graph, or another storage structure. Agents can achieve memory sharing by querying and updating this knowledge base. MetaGPT \cite{hong2023metagpt} offers a global memory pool to store all collaborative records, enabling each agent to subscribe to or search for the required information. This design allows agents to observe and extract pertinent information actively.
    \item \textbf{Shared Parameters}: In certain instances, allowing partial or complete sharing of model parameters among agents in an LLM-based MAS system may be considered. In this manner, when one agent acquires new knowledge or skills, other agents can also immediately obtain this information. However, this method may give rise to overfitting or overspecialization problems. To address this issue, the weights of shared parameters can be dynamically adjusted to balance each agent's specialization and generalization capabilities.
\end{itemize}

\subsubsection{Methods of Enhancing Communication Efficiency}

In an LLM-based MAS, the challenges of ineffective communication and LLM illusions are indeed possible. To mitigate these issues, the following strategies can be employed:

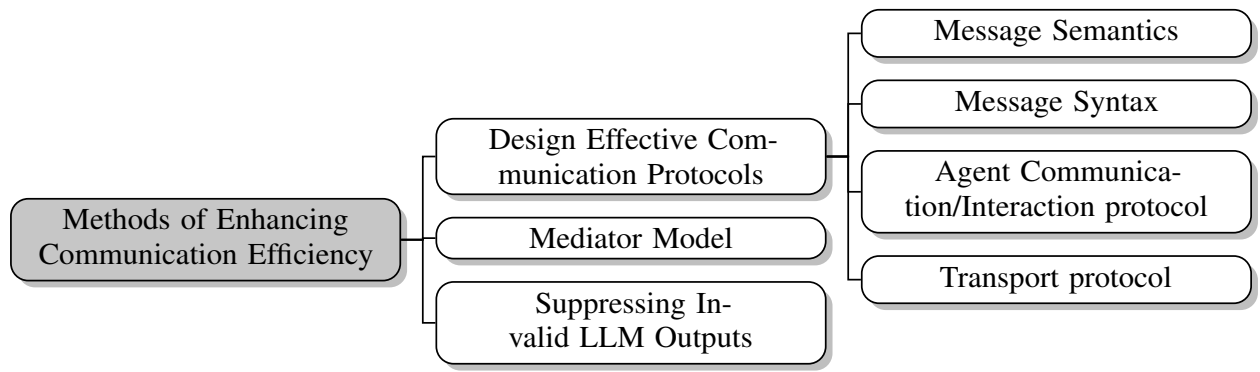
\begin{figure*}[!ht]
\scriptsize
\begin{adjustbox}{width=\textwidth}
\begin{forest}
  for tree={
    forked edges,
    grow'=0,
    draw,
    rounded corners,
    node options={align=center},
    text width=3cm,
    s sep=5pt,
    calign=edge midpoint, 
    fill=white,
    drop shadow,
  }
  [Methods of Enhancing Communication Efficiency, fill=gray!45
[Design Effective Communication Protocols
[Message Semantics]
[Message Syntax]
[Agent Communication/Interaction protocol]
[Transport protocol]
]
[Mediator Model]
[Suppressing Invalid LLM Outputs]
]
\end{forest}
\end{adjustbox} 
\caption{Typology of the Enhancing Communication Efficiency Methods.}
\label{fig:action_mindmap}
\end{figure*}

\paragraph{Design Effective Communication Protocols}

In the realm of MAS, it is imperative to scrutinize the discourse on communication through a tripartite lens that encompasses the when,' what,' and `how' dimensions. These dimensions collectively dictate the timing, content, and modality of interactions among agents, thereby serving as pivotal factors in the system's efficacy for intricate problem-solving and coordinated endeavors.

Four levels of agent communication can be identified:
\begin{itemize}
\item \textbf{Message Semantics}: The meaning of each message.
\item \textbf{Message Syntax:} The expression of each message.
\item \textbf{Agent Communication/Interaction Protocol}: The structure of conversations/dialogues.
\item \textbf{Transport Protocol}: The method of sending and receiving messages by agents.
\end{itemize}
Historically, intelligent agents, particularly those based on RL, have communicated implicitly through learning. In contrast, LLM-based agents can communicate through NLP, offering humans a more transparent and explicit mode of interaction. Consequently, concerning LLM-based MAS, concerns regarding message semantics and transport protocols are obviated.

The question of message syntax directs attention to the Agent Communication Language (ACL), which is grounded in the \emph{Speech Acts Theory} put forth by \citet{searle1969speech}. Two prominent standards have emerged: the Knowledge Query and Manipulation Language (KQML) \cite{finin1994kqml} and the ACL proposed by the Foundation for Intelligent Physical Agents (FIPA)\footnote{http://www.fipa.org/}.

In 1996, FIPA developed standards for heterogeneous and interacting agents and agent-based systems. FIPA's ACL comprises 22 performatives, or communication acts, such as Inform and Request. These performatives are not isolated entities but function as integral components of a structured conversational protocol among agents. Such protocols are regulated by predefined rules that outline the sequence and timing of performative usage to achieve specific collective objectives. For instance, FIPA-ACL can construct the FIPA-Auction-English Protocol and FIPA-Auction-Dutch Protocol.

Implementing well-defined communication protocols ensures that agent interactions adhere to a coherent structure and semantics, mitigating ambiguity and miscommunication and augmenting communication efficiency. Adopting embeddings \cite{pham2023let} or structured output formats, such as JSON, can further enhance these advantages.

\paragraph{Employing Mediator Models}
In LLM-based MAS, extensive interactions among LLMs can result in increased expenses and prolonged engagement durations. The mediator model serves as a discerning mechanism that aids in determining the necessity of interactions between LLMs, thereby reducing redundant communication overhead and enhancing the system's overall efficacy. The mediator model's decision to engage in interactions is influenced by task intricacy, the extent of inter-agent associations, and communication expenditures. Existing research has already witnessed the implementation of mediator models, with studies by \citet{huenabling, karimpanal2023lagr} delving into optimizing cost-effective, intelligent interactions between agents and LLMs.

\paragraph{Mitigating Inaccurate Outputs in LLMs}
LLMs frequently tend to generate outputs characterized by excessive praise or unfounded information. The study by \citet{wei2023simple} introduces a straightforward approach that employs synthetic data in an auxiliary fine-tuning phase to curtail the occurrence of flattering outputs. A comprehensive analysis of hallucinations in LLMs and the techniques employed to counteract them is presented by \citet{rawte2023survey}. Chain of Verification (CoVe) \cite{dhuliawala2023chainofverification} seeks to minimize hallucinations by prompting the model to initially produce a preliminary response, subsequently formulate verification inquiries for fact-checking the draft, independently address these queries, and ultimately generate a validated and refined response.

\vspace{.5em} 

By implementing these strategies, it is possible to effectively address issues of ineffective communication and LLM hallucinations in LLM-based MAS. This will ultimately enhance system performance and stability.

\section{Performance Evaluation}\label{evaluation}
\subsection{Dataset}
Most LLM-based agents do not necessitate further training of the LLM, and datasets for certain specific tasks are not publicly accessible. Consequently, we only enumerate publicly available and extensively utilized datasets.

\begin{table}[htb]
\centering
\caption{Datasets Used in the Study}
\label{tab:dataset-info-table}
\renewcommand\tabcolsep{5.2pt}
\renewcommand\arraystretch{1.2}
\resizebox{\textwidth}{!}{%
\Large
\begin{tabular}{llp{18.5cm}}

\hline
\textbf{Name} & \textbf{Field}& \textbf{Description} \\
\hline

HotpotQA \cite{yang2018hotpotqa} & Natural Language Processing & 114,000 training samples, 7,000 development set samples, and 3,000 test set samples \\
ALFWorld \cite{shridhar2020alfworld} & Symbolic Reasoning and Visual Perception & Variable, often including hundreds of environments \\
CAMEL \cite{li2023camel} & Society and Code & 50 assistant roles, 50 user roles, and 10 tasks for each combination of roles yielding  25,000 conversations for the society dataset. 20 programming languages, 50 domains, and 50 tasks for each combination of language and domains, yielding 50,000 conversations for the code dataset.\\
APPS \cite{hendrycks2021measuring} & Language Model Programming & A total of 10,000 programming questions \\
MBPP \cite{austin2021program} & Language Model Programming & 974 programming tasks \\
HumanEval \cite{chen2021evaluating} & Language Model Programming & Comprises 164 original programming problems that assess language comprehension, algorithms, and basic mathematics, with some problems being comparable to elementary software interview questions.\\
WebShop \cite{yao2022webshop} & Simulation of an e-commerce site & It contains millions of real-world products and 12,087 crowdsourced text descriptions.\\
FEVER \cite{thorne2018fever} & Natural Language Processing & Approximately 185,000 verifiable statements created by human annotators, and the evidence or rebuttals associated with those statements \\
ToolBench \cite{qin2023tool} & API use & 16,464 real-world RESTful APIs spanning 49 categories from RapidAPI Hub, and human instructions involving these APIs prompted by ChatGPT \\
MITCOURSE ES \cite{drori2022neural}&Math&25 questions from each of the seven courses: MIT’s 18.01 Single Variable Calculus, 18.02 Multivariable Calculus, 18.03 Differential Equations, 18.05 Introduction to Probability and Statistics, 18.06 Linear Algebra, 6.042 Mathematics for Computer Science, and Columbia University’s COMS3251 Computational Linear Algebra \\
RoboNet \cite{dasari2019robonet}&Robotic&An open database for sharing robotics experiences containing 15 million video frames from seven different robotics platforms\\
BridgeData V2 \cite{walke2023bridgedata}&Robotic&Robot manipulation behavior dataset containing 60,096 trajectories collected in 24 environments \\
\hline

\end{tabular}
}
\end{table}

\subsection{Benchmark}
Currently, there is no widely used benchmark for LLM-based agents, although some studies engage in comparative analysis of their LLM-based agents with others. Additionally, researchers are making strides toward proposing benchmarks that could serve as future evaluation standards.

ToolBench \cite{qin2023toolllm} is an instruction-tuning dataset for tool utilization, encompassing single-tool and multi-tool scenarios. TE \cite{aher2023using} assesses the capability of language models to emulate various facets of human behavior. \citet{akata2023playing} seeks to comprehend the social behavior of LLMs. It lays the groundwork for a behavioral game theory for machines, highlighting the substantial societal value in understanding how LLMs operate in interactive social contexts. \citet{ziems2023can} contributes a compilation of best practices for prompting and a comprehensive evaluation pipeline to gauge the zero-shot performance of 13 language models across 24 representative CSS benchmarks. 

AgentSims \cite{lin2023agentsims} offers an open-source platform for LLM evaluation. \citet{drori2022neural} involves curating a dataset of questions from the largest mathematics courses at the Massachusetts Institute of Technology (MIT) and Columbia University's Computational Linear Algebra to evaluate mathematical reasoning. BMTools \cite{qin2023tool} establishes the framework and evaluation criteria for tool utilization. SmartPlay \cite{wu2023smartplay} presents a challenging benchmark for LLM-based agents, comprising six distinct games with unique settings, offering up to 20 evaluation configurations and infinite environment variations. MLAgentBench \cite{huang2023benchmarking} is a collection of ML tasks designed for benchmarking AI research agents, facilitating operations such as reading and writing files, executing code, and examining outputs. MetaTool \cite{huang2023metatool} is utilized to evaluate whether LLMs consciously use tools and can select the appropriate ones. LLM-Co \cite{agashe2023evaluating} evaluates the ability of agents to infer cooperative partner intentions, engage in reasoning actions, and participate in long-term collaboration within a gaming environment.

\section{Prospect Applications}\label{applications}
\begin{figure}[H]
\centering
\includegraphics[width=0.5\textwidth,height=0.5\textwidth]{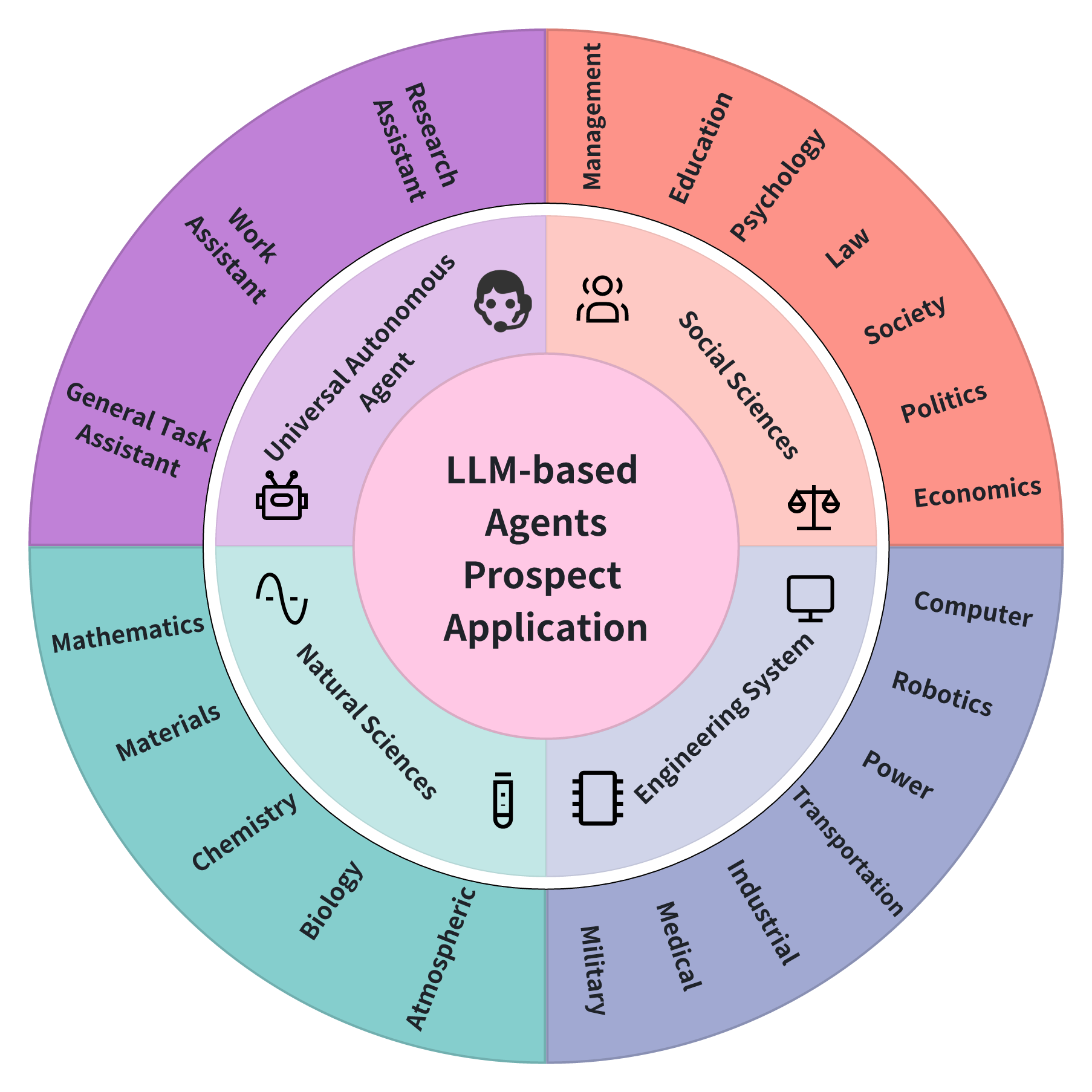}
\caption{The Prospect of LLM-based agents} \label{fig_prospect}
\end{figure}
\subsection{Natural Sciences}

\subsubsection{Mathematics}
Many recent investigations have concentrated on agents and multi-agent systems within mathematics. For example, \citet{kennedy1995particle} proposes the particle swarm optimization algorithm, a global optimization technique grounded in a multi-agent framework, which has been extensively utilized to address optimization challenges in mathematics, engineering, and computing. \citet{macal2005tutorial} discusses agent-based modeling and simulation approaches and their implementation in intricate mathematical models. \citet{crainic1986multicommodity} explores the application of agent-based methodologies in combinatorial optimization problems, specifically concerning the design of multi-commodity, multimodal transportation networks.

Currently, LLM-based agent research in mathematics predominantly emphasizes the enhancement of reasoning capabilities and support in theoretical derivation. For instance, Math Agents \cite{swan2023math} employ LLMs to investigate, uncover, resolve, and demonstrate mathematical problems. \citet{zhou2023solving} introduces an innovative and productive prompt technique, termed code-based self-verification, to further augment the mathematical reasoning potential of GPT-4's code interpreter. LeanDojo \cite{yang2023leandojo} is an instrument that can consistently interact with Lean, rectifying proof-checking errors in extant theorem-proving tools. \citet{dong2023large} conclusively determines that "P$\neq$NP" through 97 iterations of rigorous "Socratic" reasoning with GPT-4. \citet{yang2023large} devised a system capable of autonomously generating valid, original, and valuable hypotheses using only a collection of raw web texts. ToRA \cite{gou2023tora} presents a series of tool-integrated reasoning agents that utilize natural language reasoning and invoke external tools to address intricate mathematical problems. COPRA \cite{Thakur2023a} is employed for formal theorem proving, incorporating GPT-4 as a crucial element of its state-backtracking search strategy. This approach can select proof tactics throughout the search process and retrieve axioms and definitions from an external database.

LLM-based agents demonstrate substantial promise in upcoming mathematical research endeavors, encompassing:

\begin{itemize}
    \item \textbf{Aiding in Theoretical Derivation}: LLM-based agents understand prevailing theories in foundational domains, such as mathematics and physics, and facilitate human efforts in further derivation and validation, ultimately advancing scientific inquiry.
    \item \textbf{Symbolic and Numerical Computation}: LLM-based agents can be employed for symbolic and numerical computation, supporting researchers in addressing various mathematical challenges. Agents can perform numerous mathematical procedures, including solving equations, integration, differentiation, and beyond. Multi-agent systems can bolster computation expediency and precision through the collaborative partitioning of intricate mathematical problems into multiple sub-problems.
\end{itemize}

Notwithstanding the accomplishments of LLM-based agents in mathematical theory derivation and computation, it remains imperative to continually refine the mathematical reasoning capabilities of LLMs and agents and devise more efficacious mathematical knowledge representations to enhance their accuracy and efficiency in tackling complex mathematical problems. Moreover, the interpretability and dependability of LLM-based agents in resolving mathematical issues are crucial. It is vital to explore supplementary approaches to augment the interpretability of agents, empowering them to deliver more lucid and reliable solutions for users. Concurrently, supervision and validation of agent reasoning outcomes can assure their dependability in practical applications.

\subsubsection{Chemistry and Materials}
In previous research, \citet{gomez2016design} presents 1.6 million organic light-emitting diode material candidates, effectively filtered from extensive molecular libraries via high-fidelity simulations. The MolDQN framework \cite{zhou2019optimization} amalgamates chemical domain expertise with reinforcement learning methodologies to explicitly delineate molecular modifications, ensuring 100\% chemical validity. \citet{you2018graph} puts forth a Graph Convolutional Policy Network (GCPN), a model premised on a general graph convolutional network, utilized for generating goal-oriented graphs through reinforcement learning, intending to uncover novel molecules exhibiting desired attributes such as drug similarity and synthetic accessibility. \citet{beaini2023foundational} introduces the Graphium graph machine learning library, streamlining the process of constructing and training molecular machine learning models for multi-task and multi-level molecular datasets.

In the realm of current research on LLM-based agents in chemistry and materials science, Coscientist \cite{boiko2023autonomous}, leveraging the functionalities of LLM along with tools such as internet and document search, code execution, and experiment automation, is capable of autonomously designing, planning, and executing real-world chemical experiments. ChatMOF \cite{kang2023chatmof} endeavors to predict and generate Metal-Organic Frameworks (MOFs) and encompasses three core components: agents, toolkits, and evaluators. These constituents are proficient in managing data retrieval, property prediction, and structure generation. ChemCrow \cite{bran2023chemcrow} executes various chemical tasks in domains such as biosynthesis, drug discovery, and materials design by accessing chemistry-related databases, thus expediting more efficient research. LLM-based agents also demonstrate considerable potential in the following aspects:
\begin{itemize}
\item \textbf{Molecular Simulation and Chemical Reaction Optimization}: LLM-based agents can advance chemistry and materials science research by simulating molecular structures and chemical reactions. By examining various reaction pathways and conditions, these agents may pinpoint effective strategies for synthesizing novel materials or enhancing existing material properties.
    \item \textbf{Chemical Experiment Automation and Intelligence}: LLM-based agents can facilitate the automation of chemical experiments by retrieving information, querying specialized databases, and devising and implementing experimental plans tailored to particular requirements. This leads to the acquisition of data on chemical reactions and material properties. Moreover, multi-agent systems can augment the efficiency and precision of experiments via collaborative cooperation and the sharing of experimental data and experiences.
    \item \textbf{Material Design and Optimization}: In materials science research, LLM-based agents can aid in the simulation and optimization of material properties. By autonomously exploring diverse material combinations and structures and employing the robust generalization capabilities of LLMs to simulate and predict the properties of new materials, agents can uncover innovative materials with exceptional performance. This accelerates the material design process and enhances overall efficiency.
\end{itemize}
Although existing LLM-based agents have achieved some success in chemistry and materials science research, further improvement of the accuracy and reliability of models remains a significant challenge. Future research should focus on enhancing LLM's ability to handle complex chemistry and materials problems to improve the accuracy of predicting and generating chemical reactions, material properties, and other aspects.

\subsubsection{Biology}
In recent years, numerous mature studies on agents and multi-agent systems have emerged in biology. For instance, \citet{bonabeau1999swarm} explores the theory and applications of swarm intelligence, encompassing genetic algorithms, ant colony algorithms, and particle swarm algorithms based on multi-agent models. \citet{deangelis2005individual} offers a comprehensive overview of individual-based modeling methods in ecological research, simulating species interactions and environmental impacts within ecosystems. \citet{wilensky2015introduction} presents agent-based modeling methods and their applications in natural, social, and engineering complex systems, including simulating geographical science issues such as ocean ecosystems and atmospheric circulation using agent systems. \citet{jain2022biological} proposes an active learning algorithm utilizing GFlowNets as generators of diverse candidate solutions, aiming to produce biological sequences with optimal characteristics, such as protein and DNA sequences.


Presently, research on LLM-based agents in biology is limited. BioPlanner\cite{odonoghue2023bioplanner} is an automated evaluation approach for assessing the performance of LLM in protocol generation and planning tasks within the domain of biology. OceanGPT \cite{bi2023oceangpt} employs multi-agent collaboration for the automatic generation of data in various subfields of ocean science. Nevertheless, there exists substantial potential for future investigations in areas such as:

\begin{itemize}
    \item \textbf{Ecosystem Modeling}: LLM-based agents can simulate species interactions and environmental impacts within ecosystems, aiding researchers in comprehending ecosystem structure and function. For example, ecosystem stability, diversity, and evolutionary processes can be analyzed by simulating the behavior and interactions of various agents, including biological individuals, populations, and environments.
    \item \textbf{Group Behavior and Collective Intelligence}: Through the simulation of behavior and interactions within groups, fundamental concepts of group behavior, collective intelligence, population genetics, and evolution can be elucidated. In particular, by simulating the behavior and interactions of multiple agents, such as molecular or biological groups, group behavior's formation, coordination, adaptation, and evolution can be examined, leading to a better understanding of the mechanisms governing overall system functioning.
    \item \textbf{Cell Biology and Molecular Biology}: LLM-based agents can simulate molecular mechanisms and signaling pathways within cells, subsequently investigating interactions and regulation between biomolecules. For instance, biological processes like intracellular signal transduction, gene expression regulation, and metabolic pathways can be analyzed by simulating the behavior and interactions of multiple agents, such as proteins, nucleic acids, and metabolites.
\end{itemize}

Biological systems are known for their inherent complexity, manifesting across various hierarchical levels, spatiotemporal scales, and temporal extents. In light of this, agents utilizing LLMs must demonstrate proficiency in managing this intricacy. This entails accounting for diverse biological entities' dynamic behaviors and interactions, including individual organisms, populations, and their respective ecological contexts. Furthermore, data within the realm of biology often exhibits attributes of being voluminous, diverse, heterogeneously structured, and subject to inherent noise. This is evident in datasets encompassing genomic, phenotypic, and environmental information. Consequently, LLM-based agents are required to possess the capability to effectively process substantial volumes of heterogeneous data and distill valuable insights and knowledge from it.

\subsubsection{Climate Science}
In atmospheric research, the employment of agent systems has predominantly spanned areas such as the elucidation of climate behavior and the investigation of climate energy economics. A novel agent-based modeling methodology is presented by \citet{jager2021using}, which expounds its utility in deciphering the intricacies of climate-related behavioral dynamics. Furthermore, \citet{castro2020review} offers a comprehensive examination of studies centered on climate-energy policies, emphasizing the reduction of emissions and energy conservation by implementing agent-based modeling approaches.

In the current research landscape, \citet{kraus2023enhancing} leverages LLM-based agents to extract emission data from ClimateWatch\footnote{https://www.climatewatchdata.org/}, thereby furnishing more accurate and dependable data pertinent to crucial facets of climate change. LLM-based agents can be harnessed for climate change prognostication by deploying sensor networks across diverse geographical locales to collect atmospheric data (e.g., temperature, pressure, humidity, wind speed) and performing real-time analysis and processing of this data via LLM-based agents. This methodology can further anticipate or issue alerts for atmospheric phenomena and climate change. In contrast, within the realm of climate model simulation and optimization, LLM-based agents can emulate various atmospheric processes and events, such as atmospheric circulation, climate systems, and the propagation of air pollution. By perpetually optimizing and modifying the interaction rules among agents, the model can be honed to more accurately reflect real-world scenarios, ultimately producing more precise predictions and solutions for atmospheric science research.
During the climate simulation process, the escalating complexity of MAS presents a substantial challenge to computational efficiency. Enhancing the performance of LLM-based agents' planning and reconsideration is vital for achieving more precise climate simulations and forecasts within limited computational resources. Moreover, since most atmospheric data are numerical, enhancing LLM's understanding and computational capacity for numeric values will significantly influence system performance.

\subsection{Universal Autonomous Agent} 
\subsubsection{General Task Assistant} 
Current research on general task assistants primarily focuses on the LLM-based agent systems or frameworks. A Generalist Agent \cite{reed2022generalist} is a multimodal, multi-task, and multi-entity universal agent capable of performing various tasks, such as playing Atari games, naming images, chatting, stacking blocks with real robotic arms, and more. HuggingGPT \cite{shen2023hugginggpt} integrates various modules and AI models from different domains within the machine learning community to execute task planning. ModelScope-Agent \cite{li2023modelscope} is a universal, customizable LLM-based agent framework for practical applications, providing a user-friendly system library. LangChain \cite{githubGitHubLangchainailangchain} is an open-source framework that enables efficient software development through natural language communication and collaboration. XLang \cite{githubGitHubXlangaixlang} offers a comprehensive set of tools and user interfaces for LLM-based agents, supporting data processing, plugin usage, and web scenarios. BabyAGI \cite{githubGitHubYoheinakajimababyagi} creates tasks based on predefined objectives, utilizes LLM to create new tasks, and stores and retrieves task results. AutoGPT \cite{githubGitHubSignificantGravitasAutoGPT} is an automated agent capable of breaking down objectives and executing tasks in a loop. AgentVerse \cite{chen2023agentverse} enables the rapid creation of simulation experiments based on multiple LLM-based agents performing different roles. LMA3 \cite{colas2023augmenting} is a method that leverages LLM to support various abstract objective representations, generation, and learning. Kani \cite{zhu2023kani} assists developers in implementing various complex functionalities by providing core building blocks for chat-based interactions, including model interfaces, chat management, and powerful function invocation.

Although notable advancements have been made in the study of general task assistants employing LLM-based agents, several challenges persist. One such challenge is determining how to judiciously control the granularity of task decomposition while maintaining task-solving efficiency, minimizing token consumption, and reducing computational resource demands. Another challenge pertains to memory utilization and information integration: devising methods to more effectively employ information stored in memory, amalgamate knowledge and data from disparate sources, and augment the accuracy and efficiency of LLM-based agents in problem-solving. Furthermore, developing additional tools and techniques to bolster LLM-based agents in various contexts is essential, enhancing their adaptability and scalability in general task assistants. Ultimately, equipping LLM-based agents with long-term learning and adaptive capabilities is crucial for continuously ameliorating performance in the face of ever-evolving tasks and environments.

Future research may investigate more efficient automated task decomposition and optimization algorithms, enabling LLM-based agents to autonomously execute reasonable task decomposition when confronted with complex tasks, thereby improving problem-solving speed and quality. Additionally, integrating multimodal information processing techniques into agents will facilitate the handling and integration of information from different modalities, such as images, sounds, and videos, thereby enriching the capabilities of task assistants.

\subsubsection{Work/Research Assistant} In the context of work and research endeavors, the accumulation of substantial volumes of materials and literature may be requisite, followed by their summarization through comprehension, the refinement of viewpoints post-experimentation and validation, and ultimately, their compilation into reports, papers, presentations, or narrative and cinematic works. These steps can also be entrusted to LLM-based agents, which can browse web pages, databases, and literature repositories, summarize them via LLM, generate experimental code for validation, and subsequently draft conclusions.

In general text generation, ChatEval \cite{chan2023chateval} employs a multi-agent debate framework, enhancing efficiency and effectiveness in handling complex tasks. \citet{zhu2023towards} proposes a heuristic reinforcement learning framework that can significantly improve performance without requiring preference data. In creating research reports, stories, and television dramas, \citet{fablestudioSHOW1Showrunner} presents a method based on LLMs, custom diffusion models, and multi-agent simulations to generate high-quality episodic content. GPT Researcher \cite{githubGitHubAssafelovicgptresearcher} is an autonomous agent capable of producing detailed, accurate, and unbiased research reports. \citet{boiko2023emergent} proposes an intelligent agent system capable of autonomously designing, planning, and executing complex scientific experiments. In domain-specific applications, \citet{mehta2023improving} constructs an agent that learns to understand architect language instructions and uses them to place blocks on a grid, aiming to build a 3D structure. LayoutGPT \cite{feng2023layoutgpt} collaborates with visual generative models to produce reasonable layouts in various domains, from 2D images to 3D indoor scenes. MusicAgent \cite{yu2023musicagent} incorporates music-related tools and autonomous workflows, including timbre synthesis and music classification, to address user requirements. MemWalker \cite{chen2023walking} is an interactive agent designed for long-text reading, which utilizes a technique to convert extensive contexts into a tree structure of summary nodes. When a query is received, the agent traverses this tree to locate pertinent information and generates a response after accumulating adequate information.

In contrast to general task assistants, work and research assistants demand more robust memory and knowledge integration capabilities. Augmenting the memory capacity of LLM-based agents is vital for efficiently organizing, summarizing, and retrieving information after processing extensive volumes of textual material. Furthermore, effectively utilizing domain-specific tools, such as code and simulators, for validation experiments is crucial for enhancing task completion and accuracy. LLM-based agents should also exhibit a more comprehensive array of cross-domain knowledge and skills to adapt to diverse work and research requirements. Ultimately, innovation and originality pose significant challenges, as it is necessary to bolster the creativity and originality of LLM-based agents in work and research assistance while circumventing the generation of repetitive or excessively similar content.

Future endeavors in LLM-based agents' work and research assistance may delve further into domains such as artistic creation in music and film generation and incorporate human-machine collaboration to capitalize on human knowledge for producing more original works, thereby providing greater convenience to human work and creativity.

\subsection{Social Sciences}
\subsubsection{Economics and Finance } Existing agents and multi-agent systems have been applied in economic and financial research. \citet{arthur2018asset} employs a multi-agent model to construct an artificial stock market, exploring financial market issues such as asset pricing, investor behavior, and market volatility. \citet{tesfatsion2006handbook} comprehensively introduces agent-based computational economics methods and their applications in various economic fields. \citet{johanson2022emergent} demonstrates agents' ability to generate resources in space using MARL and trade them at their preferred prices. The AI Economist \cite{zheng2020ai} proposes an economic simulation environment with competitive pressures and market dynamics, validating the simulation by demonstrating the operation of basic taxation systems in an economically consistent manner, including the behavior and specialization of learning and professional agents. \citet{tilbury2022reinforcement} reviews the historical barriers classical agent-based techniques face in economic modeling. The AI Economist: Improving Equality and Productivity with AI-Driven Tax Policies presents a two-level deep reinforcement learning method based on economic simulations for learning dynamic tax policies, with agents and governments learning and adapting.

Currently, many studies focus on LLM-based agents in economics and finance. \citet{horton2023large} compares LLM behavior with actual human behavior by placing LLMs in different economic scenarios and exploring their behavior. This enables researchers to investigate economic behavior in simulations such as dictator games and minimum wage issues, gaining new insights into economics. \citet{phelps2023models} investigates LLM responses in principal-agent conflicts, with LLM-based agents overriding their principal's objectives in a simple online shopping task, providing clear evidence of principal-agent conflict and highlighting the importance of incorporating economic principles into the alignment process. AucArena \cite{chen2023money} illustrates the efficient involvement of LLM-based agents in auctions, effectively managing budgets, preserving long-term objectives, and improving adaptability through explicit incentivization mechanisms. 

In the domain of Game Theory, the Suspicion-Agent \cite{guo2023suspicionagent} exhibits exceptional adaptability in various imperfect information card games. It demonstrates robust higher-order Theory of Mind capabilities, suggesting it can comprehend others and intentionally influence their behavior.  

Numerous studies have investigated using LLM-based agents in the context of financial transaction scenarios. AlphaGPT \cite{wang2023alphagpt} introduces an interactive framework for Alpha mining, which employs a heuristic method to comprehend the concepts utilized by quantitative researchers, subsequently generating innovative, insightful, and efficient Alphas. TradingGPT \cite{li2023tradinggpt} presents a novel LLM-based MAS framework with layered memories to enhance financial trading decisions by simulating human cognitive processes. This approach enables agents to prioritize crucial tasks, integrate historical actions and market insights, and engage in inter-agent discussions, improving responsiveness and accuracy.

Given LLM-based agents' enhanced textual comprehension and complex decision-making capabilities, there is considerable potential for research in economics and finance utilizing these agents. Relevant explorations may encompass the following areas:

\begin{itemize}
\item \textbf{Market Simulation and Emulation}: Establishing LLM-based agents to simulate the behavior of various market actors, such as supply and demand sides, competitors, and regulators, can enable researchers to predict and emulate data on product prices, market shares, market structures, and transaction completion rates. Behaviors may include purchasing, competitive bidding, bargaining, and collaborative tendering.
\item \textbf{Financial Market Analysis}: Through the simulation of actions by financial market participants, including investors, institutions, and regulators, LLM-based agents can offer valuable insights into market volatility and risks. For instance, simulations of investor trading behavior and market information dissemination processes can provide predictions about fluctuations in stock prices, exchange rates, and interest rates.
\item \textbf{Macroeconomic and Policy Simulation}: LLM-based agents can model the implementation process of fiscal and monetary policies, incorporating various economic actors such as governments, businesses, and individuals. This allows the agents to forecast shifts in macroeconomic indicators, including GDP, inflation, and unemployment rates.
\item \textbf{Socio-economic Network Analysis}: By modeling processes such as information dissemination, resource allocation, and trust-building within socio-economic networks, LLM-based agents can contribute to a more profound understanding of the evolution and impact of network economies. Specifically, simulations involving diverse agents, such as consumers, businesses, and governments, can offer insights into network effects, information asymmetry, and market failures.
\end{itemize}
For LLM-based agents in economics, typically simulating human or economic actors' decision-making, the action space of agent interactions and the state of the agent plays a crucial role, directly affecting experimental results. Effectively representing the interaction action space and the state of the agent to simulate the decision-making process of economic actors more accurately is a significant challenge. Simultaneously, the credibility of LLM anthropomorphism is also a major challenge. If conducting large-scale macroeconomic analysis, many LLM-based agents may be required, posing difficulties for system performance or token consumption. One approach is to use reinforcement learning methods to control and reduce the number of interactions with LLMs.

\subsubsection{Politics} In previous agent research within the political domain, \citet{epstein1996growing} employs MAS to construct an artificial society, investigating the formation and evolution of social phenomena, including political communication and social movements in political science. \citet{lustick2009abstractions} discusses the application of MAS in comparative political science research, encompassing political systems, political decision-making, and political stability. \citet{tsvetovat2004modeling} introduces the application of multi-agent models in studying complex socio-technical systems, including political communication and political decision-making in political science. \citet{trott2021building} utilizes two-level RL and data-driven simulation to achieve effective, flexible, and interpretable policy design.

In current LLM-based agents research, these agents are employed to explore potential decisions and communication situations of political actors. \citet{ziems2023can} uses LLM-based agents to help understand the content and tactics of politicians' speeches. \citet{bail2023can} demonstrates that LLM-based agents can detect ideology, predict voting results, and identify patterns. \citet{mukobi2023welfare} presents a general-sum variant of the zero-sum board game diplomacy. In this variant, agents must balance investments in military conquest and domestic welfare.

LLM-based agents in the political domain can explore the following areas:
\begin{itemize}
\item \textbf{Political Simulation and Prediction}: By simulating the behaviors and interactions of various participants in the political process, such as party competition, voter behavior, and policy-making processes, LLM-based agents can forecast developmental trends of political events, election outcomes, and policy effects.
\item \textbf{Political Decision-making Analysis}: Employing LLM-based agents to simulate the behaviors and interactions of different political decision-making processes enables the evaluation of the advantages, disadvantages, and impacts of various policy choices. This approach allows researchers to simulate interactions among governments, political parties, and interest groups, providing policymakers with valuable information on policy effects.
\item \textbf{International Relations Research}: Utilizing LLM-based agents to model interactions and conflicts between countries in international politics, researchers can explore various aspects such as international trade, military conflicts, and diplomatic interactions. This approach assists in understanding the complexity and potential risks associated with international politics.
\end{itemize}

In political research, LLM-based agents may need to ensure communication efficiency while avoiding excessive politeness and ineffective communication, enhancing the practical application value of LLM-based agents in political science research. Simultaneously, accurately modeling the complexity and uncertainty of political environments to improve the accuracy and reliability of LLM-based agents in political domain research poses a challenge. Of course, it is also necessary to ensure that LLM-based agent behavior complies with ethical and moral requirements, avoiding adverse social impacts.

\subsubsection{Society}
In prior Multi-Agent research in sociology, \citet{epstein1996growing} employs multi-agent models to construct an artificial society, investigating the formation and evolution of social phenomena such as social movements, cultural evolution, and social change within the field of sociology. \citet{macy2002factors} introduces computational sociology and agent-based modeling methods, encompassing social networks, social norms, and social influence within sociology. \citet{gilbert2005simulation} presented the theory and practice of utilizing simulation methods for social scientists, including the application of multi-agent models in sociological research. \citet{hasan2023artificial} discusses the pillars of sustainable development (e.g., social, environmental, and economic).

Currently, LLM-based agents primarily concentrate on simulating human behavior and social interactions. Generative Agents \cite{park2023generative} proposes an interaction mode of Multi LLM-based agents to achieve a credible simulation of human behavior. \citet{gao2023s} uses prompt engineering and adjustment techniques to create an LLM-based MAS that simulates real-world social network data, including emotions, attitudes, and interaction behaviors. \citet{li2023you} examine the behavior characteristics of LLM-driven social robots within a Twitter-like social network. The results demonstrated that these robots could disguise and influence online communities through toxic behavior. \citet{liu2023training} proposes a novel learning paradigm enabling language models to learn from simulated social interactions. \citet{feng2023role} investigates the capability of LLM-based agents to simulate credible human behavior in carefully designed environments and protocols. \citet{wei2023multi} assesses the performance of multi-party group chat conversation models, explores methods to enhance model performance, and addresses the challenges of turn-taking and dialogue coherence. 

On the other hand, \citet{li2023quantifying} develops an opinion network dynamic model to encode LLM opinions, individual cognitive acceptability, and usage strategies, simulating the impact of LLMs on opinion dynamics in various scenarios. LLM-Mob \cite{wang2023would} utilize LLM's language understanding and reasoning capabilities to analyze human migration data by introducing the concepts of historical stays and contextual stays, capturing long-term and short-term dependencies of human movement, and employing the temporal information of prediction targets for time-aware prediction. \citet{egami2023using} utilizes LLM outputs for downstream statistical analysis of document labels in social science while maintaining statistical properties, such as asymptotic unbiasedness and accurate uncertainty quantification. \citet{ghaffarzadegan2023generative} explores the emerging opportunities in employing generative artificial intelligence for constructing computational models with intricate feedback, which can depict individual decision-making within social systems. Lyfe Agents \cite{kaiya2023lyfe} assesses the self-motivation and social capabilities of the agents in various multi-agent scenarios. The approach combines low-cost and real-time responsiveness while preserving intelligence and goal-directedness.

These studies offer various methods and frameworks for LLM-based agents in simulating human behavior and social interactions. Owing to LLM-based agents' ability to simulate human communication and mimic human thinking, these agents can simulate credible human behavior, participate in multi-party group chats, learn social interactions in simulated environments, handle memory and planning tasks, and exhibit human behavior characteristics in opinion dynamics.

However, these studies also reveal challenges, such as ensuring that LLM-based agents maintain turn-taking and dialogue coherence in multi-party group chats to enhance the authenticity of simulating human behavior and social interactions and effectively training socially-aligned language models in simulated environments to improve LLM-based agents' adaptability and accuracy in social interactions. Additionally, LLM-based agents must achieve diversity and personalized simulation for each human actor to better reflect real-world social phenomena. Future research may continue to explore these challenges and propose more effective methods to improve LLM-based agents' performance in simulating human behavior and social interactions.

\subsubsection{Law} In prior research on Agent and Multi-Agent systems within the legal domain, \citet{bench2003model} employs multi-agent models to examine theories and values in the legal reasoning process, offering novel theories and methods for legal decision-making and legal system design. \citet{branting2013reasoning} constructs a computational legal analysis model using multi-agent models, investigating the role of legal rules and precedents in legal reasoning.

Currently, there is limited research on LLM-based agents in the legal domain. Blind Judgement \cite{hamilton2023blind} introduces Multi-LLM-based agents for simulating judicial decisions of the United States Supreme Court from 2010 to 2016, training nine separate models to emulate the opinions of different justices. \citet{shui2023comprehensive} assesses the efficacy of LLMs when integrated with professional information retrieval systems for case-based learning and question-answering within the legal field.

Considering that LLM-based agents possess robust text processing and comprehension capabilities, as well as a Memory mechanism for recording historical cases and decisions, there is substantial potential for exploration in the legal domain, such as in the following areas:
\begin{itemize}
\item \textbf{Autonomous Legal Assistant}: LLM-based agents integrate legal provisions and historical case reviews to provide document writing and auxiliary advice on current cases. 
\item \textbf{Legal Decision-making Analysis}: LLM-based agents simulate the behavior and interactions of various participants in the legal decision-making process, including judges, lawyers, and litigants, to evaluate the advantages, disadvantages, impacts, fairness, and efficiency of distinct legal policies and legal systems.
\end{itemize}
As the legal domain typically involves substantial textual material, the LLM in LLM-based agents requires a longer context and more efficient Memory capabilities. Furthermore, effectively representing legal knowledge, encompassing legal provisions, historical cases, and legal principles, and executing accurate legal reasoning in LLM-based agents are crucial for making decisions or simulating after reading and comprehending the law.

\subsubsection{Psychology} In previous research in psychology, \citet{sun2006cognition} provides a comprehensive introduction to applying multi-agent interaction in cognitive modeling and social simulation, encompassing cognitive processes, social interactions, and emotional motivations in psychology. \citet{marsella2009ema} employs agent models to model the emotional appraisal process, enabling a deeper understanding of the fundamental principles of emotional psychology.

Currently, LLM-based agents primarily focus on applications in mental health support and psychological experiment simulation. \citet{ma2023understanding} conducts a qualitative analysis of LLM-based agent-supported mental health support applications. The study finds that the application helps provide on-demand, non-judgmental support, enhancing user confidence and facilitating self-discovery. However, it faces challenges in filtering harmful content, maintaining consistent communication, remembering new information, and alleviating user overdependence. \citet{aher2023using} utilizes LLM-based agents to simulate psychological experiments, revealing some "hyper-precise distortions" in LLM that could affect downstream applications. \citet{akata2023playing} employs LLM-based agents to simulate repeated games in game theory, discovering that LLM-based agents perform exceptionally well in games emphasizing self-interest, especially in prisoner's dilemma games, and exhibit a psychological tendency to prioritize self-interest over coordination. These studies provide various methods and frameworks for LLM-based agents in mental health support and psychological experiment simulation. These LLM-based agents have broad application prospects in providing psychological support, replicating psychological findings, and simulating game theory experiments. Humanoid agents \cite{wang2023humanoid} constitute a platform for developing agents that emulate human cognition, communication, and behavioral patterns. These agents incorporate logical reasoning capabilities contingent on particular factors, such as fulfilling fundamental needs, emotions, and interactions with others. \citet{zhang2023exploring} investigates the potential of LLM-based multi-agent societies to mirror human collaborative intelligence.

LLM-based agents in the field of psychology can explore the following areas in the future:
\begin{itemize}
\item \textbf{Psychological Therapy and Counseling}: By simulating interactions and influences during psychological therapy and counseling processes, LLM-based agents contribute to a deeper understanding of the fundamental principles of psychotherapy and counseling psychology for researchers and support patients receiving psychological treatment. 
\item \textbf{Cognitive Modeling}: Through emulating cognitive processes such as perception, memory, thinking, and decision-making, LLM-based agents offer insights into the core principles of cognitive psychology. Specifically, these agents can analyze cognitive biases and strategies by simulating individuals' cognitive processes across various situations. 
\item \textbf{Emotion and Motivation Modeling}: Utilizing LLM and Memory to model emotional and motivational processes, LLM-based agents enable researchers to explore the fundamental principles of emotional and motivational psychology by examining emotional reactions, interests, and drives in individuals.
\end{itemize}
However, these studies also reveal challenges, such as effectively filtering harmful content, maintaining consistent communication, achieving more anthropomorphic communication or simulation, and addressing user overdependence issues. Future research may continue to explore these challenges and propose more effective methods to improve the performance of LLM-based agents in mental health support and psychological experiment simulation.

\subsubsection{Education} In existing Agent and Multi-Agent research, \citet{woolf2010building} introduces methods and techniques for constructing intelligent interactive tutors, including using Agent and Multi-Agent systems to implement personalized teaching and adaptive learning. \citet{soller2003computational} presents computational methods employing multi-agent models to analyze online knowledge-sharing interactions to improve educational organization and management.

Owing to their strong natural language interaction capabilities, LLM-based agents facilitate efficient communication with humans, which can be useful for assisting human learning or simulating classrooms in the education domain. For research assistance, please refer to the research assistance section. Math Agents \cite{swan2023math} convert mathematical formulas from literature into LaTeX and Python formats, utilizing LLM as a language user interface and artificial intelligence assistant to foster interaction between mathematics and computer science. AgentVerse \cite{chen2023agentverse} is an LLM-based MAS framework simulating NLP classroom education. CGMI \cite{jinxin2023cgmi} is a general multi-agent interaction framework that simulates various classroom interactions between teachers and students, with experimental results demonstrating that teaching methods, courses, and student performance closely resemble real classroom environments. Furthermore, LLM-based agents can simulate the implementation process of future educational policies and systems, assisting researchers in evaluating the advantages, disadvantages, and impacts of different educational strategies. For instance, by simulating the behavior of governments, schools, teachers, and students, MAS can predict academic input, quality, and equity changes.

In the education domain, the primary challenge for LLM-based agents is to output harmless, more credible content to enhance education quality. Another challenge is diversity and personalization: education targets a diverse range of students, and implementing personalized teaching and adaptive learning for each student in the LLM-based agent system remains a significant challenge. Moreover, although LLMs possess strong natural language interaction capabilities, there is room for improvement in understanding students' questions, expressions, and emotions to address their learning needs better.

\subsubsection{Management} In the existing research on Agent and Multi-Agent Systems domains, \citet{north2007managing} provides a comprehensive introduction to applying agent-based modeling and simulation for managing business complexity. This encompasses organizational behavior, human resource management, and marketing in management studies. \citet{bonabeau2002agent} introduces agent-based modeling methods and their applications in simulating human systems, including organizational behavior, supply chain management, and financial markets in the management field. \citet{liu2022multi} applies MARL to multi-echelon inventory management problems, aiming to minimize the overall supply chain costs.

Currently, LLM-based agents in the management domain primarily focus on simulating the operations of companies and organizations. For instance, MetaGPT \cite{hong2023metagpt} and ChatDev \cite{qian2023communicative} simulate multiple roles in a software company for collaborative software development. MetaAgents \cite{li2023metaagents} utilizes a simulated job fair environment as a case study to assess agents' information processing, retrieval, and coordination capabilities. The results indicate that these agents demonstrate exceptional performance in comprehending project workflows, identifying appropriate co-authors, and delegating tasks. Further exploration can be conducted in the following areas:
\begin{itemize}
\item \textbf{Organizational Behavior and Collaborative Work}: By simulating the behavior and interactions of employees, teams, and managers in organizations, LLM-based agents serve as a valuable tool for researchers to study collaborative work processes, enhancing the understanding of organizational structure, culture, leadership, and efficiency.
\item \textbf{Company Auxiliary Operations}: Assisting companies and organizations in their operations, LLM-based agents contribute to increased efficiency by taking on reporting, information summarization, processing, approvals, and decision-making, resulting in more efficient, fair, and transparent company operations.
\item \textbf{Supply Chain Management and Logistics Optimization}: Researchers can effectively analyze and optimize supply chain management and logistics by employing LLM-based agents to simulate resource allocation and collaborative decision-making processes. This is achieved by modeling the behavior and interactions of suppliers, manufacturers, distributors, and retailers, allowing LLM-based agents to address supply chain inventory management, transportation schedules, and demand forecasting issues.
\end{itemize}
Management problems often involve multiple levels, roles, and objectives. Effectively addressing these complexities and scalability issues in LLM-based agents remains a significant challenge. Furthermore, management research typically relies on various historical and real-time data forms. LLM-based agents need to understand historical data in different formats effectively.

\subsection{Engineering Systems}
\subsubsection{Computer System}

In computer science, numerous mature studies exist on agent and multi-agent systems. These studies primarily focus on computer operation tasks, human-computer interaction, code generation and testing, network security, gaming, and recommendation system applications. 

\begin{itemize}
    \item \textbf{Computer Operation}: RCI \cite{kim2023language} employs natural language commands to guide LLMs in completing computer tasks. Mobile-Env \cite{zhang2023mobileenv} is based on the Android for mobile devices environment, enabling intelligent agents to observe Android operating system screenshots, view hierarchies, and interact with Android applications.
    \item \textbf{Human-computer Interaction}:  \citet{lin2023decision} introduces a collaborative task called decision-oriented dialogue. In these tasks, AI assistants collaborate with humans through natural language to assist in making complex decisions. SAPIEN \cite{hasan2023sapien} introduces a high-fidelity virtual agent platform driven by LLMs, allowing open-domain conversations with users in 13 languages and expressing emotions through facial expressions and voice modulation. In web interaction, WebAgent \cite{gur2023real} proposes a model that integrates two language models—a domain expert language model and a general language model—for autonomous navigation on real websites. WebArena \cite{zhou2023webarena} is a standalone, self-hosted web environment for building autonomous agents. SheetCopilot \cite{li2023sheetcopilot} facilitates interaction with spreadsheets using natural language, converting complex requests into actionable steps.
    \item \textbf{Network Security}: \citet{rigaki2023out} proposes a method that uses LLM as attack agents, applied in reinforcement learning environments. 
    \item \textbf{Code Generation}: GPT-Engineer \cite{githubGitHubAntonOsikagptengineer} is easy to adapt and extend, allowing LLM-based agents to generate entire code repositories based on prompts. \citet{dong2023self} allows multiple LLMs to play different roles, forming a team without human intervention to collaborate on code generation tasks. ChatDev \cite{qian2023communicative} explores using LLM-driven end-to-end software development frameworks, covering requirements analysis, code development, system testing, and document generation to provide a unified, efficient, cost-effective software development paradigm. CAAFE \cite{hollmann2023llms} employs LLMs to generate and execute code for feature engineering on tabular datasets. AutoGen \cite{wu2023autogen} presents an autonomous LLM-based agent that generates an entire code repository based on prompts.
    \item \textbf{Software Testing}: LLift \cite{li2023hitchhikers} is an interface with static analysis tools and LLM, using carefully designed agents and prompts for full automation. \citet{feldt2023towards}  proposes an autonomous LLM-based testing agent that provides a conversational framework to help developers with testing and emphasizes the benefits of LLM illusions in testing. RCAgent \cite{wang2023rcagent} is a tool-enhanced agent for practical and privacy-aware industrial root cause analysis (RCA) in cloud environments.
    \item \textbf{Recommendation System}: RecAgent \cite{wang2023recagent} uses LLM as the brain and recommendation models as tools, creating a versatile and interactive recommendation system. Agent4Rec \cite{zhang2023generative} comprises user profile, memory, and action modules and interacts via web pages to deliver personalized movie recommendations.
    \item \textbf{Role-playing Game}: VOYAGER \cite{wang2023voyager} is a lifelong learning agent in Minecraft driven by LLM that continuously explores the world, acquires various skills, and makes discoveries. GITM \cite{zhu2023ghost} proposes a framework that achieves efficient and flexible operation by converting long-term and complex goals into a series of lowest-level keyboard and mouse operations. \citet{junprung2023exploring} proposes two agents simulating human behavior: a two-agent negotiation and a six-agent murder mystery game. \citet{zhou2023dialogue} proposes a dialogue-shaping framework that allows LLM to obtain helpful information from NPCs through dialogue and convert it into a knowledge graph, then use story-shaping techniques to accelerate RL agents' convergence to optimal strategies. Clembench \cite{chalamalasetti2023clembench} has developed a flexible and scalable framework using conversational games as testing tools to assess a wide range of models quickly. Tachikuma \cite{liang2023tachikuma} proposes the integration of virtual game masters (GMs) into the world models of agents. GMs play a crucial role in supervising information, estimating player intentions, providing environment descriptions, offering feedback, and addressing the limitations of current world models.  \citet{xu2023exploring} effectively conducts the Werewolf game without adjusting the LLMs' parameters and exhibits strategic behavior in experiments. MindAgent \cite{gong2023mindagent} presents a novel game scenario and associated benchmarks, facilitating the evaluation of multi-agent collaboration efficiency and enabling the supervision of multiple agents engaged in gameplay concurrently. 
    \item \textbf{Game Generation}: \citet{chen2023ambient} designs a text-based adventure game imaginative play system that generates stories related to imaginative play based on ChatGPT. GameGPT \cite{chen2023gamegpt} utilizes a dual-agent collaboration and a hierarchical approach, employing multiple internal dictionaries for automating game development.
\end{itemize}

Despite some achievements, many research directions and challenges exist for LLM-based agents in computer science. For example, in code generation and testing, the coding capabilities of LLM are essential, and how to improve the code quality and testing results of LLM-based agents is a noteworthy issue. In network security, recommendation systems, and other aspects, fully utilizing the advantages of LLM-based agents and addressing existing problems still require further research. For computer operation and human-computer interaction, LLM-based agents must master more tool usage capabilities to implement more functions. Further, by building adaptive learning and long-term development LLM-based agent systems, they can continuously improve their performance when facing constantly changing computer science problems.

\subsubsection{Robotics System}
In previous research on agents and multi-agent systems in robotics, \citet{parker2016multiple} introduced investigations on multi-mobile robot systems and collaborative control issues among multiple robots. \citet{busoniu2008comprehensive} discussed robotic learning and intelligence concerns.

In the present work on LLM-based agent research within robotics, the primary focus lies on robot task planning. \citet{di2023towards} proposes a framework that utilizes language as a core reasoning tool, simulates robot operation environments, and demonstrates significant performance improvements in exploration efficiency and offline data reuse.  ProgPrompt \cite{singh2022progprompt} proposes a programmatic LLM prompt structure that facilitates task planning across various environments and robotic functional tasks. \citet{huang2022inner} examines how LLM can execute reasoning via natural language feedback in robotic control situations without requiring further training. TaPA \cite{wu2023embodied} presents a method for planning in the real world under physical scene constraints, where agents generate executable plans by aligning LLM and visual perception models based on the objects in the scene. LLM-Planner \cite{song2022llm} leverages the power of LLMs for sample-efficient planning for embodied agents. \citet{xiang2023language} fine-tunes LLM with world models to acquire diverse embodied knowledge, using these experiences to fine-tune LLM further and enable reasoning and action in various physical environments. 3D-LLM \cite{hong20233d} accepts 3D point clouds and their features as input, completing a series of 3D-related tasks. ProAgent \cite{zhang2023proagent} can predict teammates' upcoming decisions and develop enhanced plans for itself, demonstrating exceptional performance in cooperative reasoning. Additionally, it can dynamically adjust its behavior to improve collaboration with teammates.

LLM-based agents hold promising potential in enhancing automation levels, supporting multi-scenario applications, and achieving efficient task execution. Future research may continue to address these challenges or investigate the following aspects:
\begin{itemize}
\item \textbf{Multi-robot Collaborative Control}: LLM-based agents are well-suited for simulating collaborative control and task allocation in multi-robot systems, assisting researchers in improving the collaborative performance and execution efficiency of such systems. For instance, researchers can analyze multi-robot task allocation, path planning, and collaboration strategies by simulating the behavior and interactions of various types of robots, tasks, and environments.
\item \textbf{Unmanned Aerial Vehicle (UAV) Swarm Flight and Control}: LLM-based agents can simulate swarm control, path planning, and obstacle avoidance in UAV swarm flight, aiding researchers in analyzing flight stability, formation change, and safe flight of UAV swarms.
\end{itemize}
Concurrently, LLM-based agents must address complex environment adaptation and modeling more comprehensively, as robotics encompasses numerous complex environments and tasks that demand accurate handling of complex issues. Moreover, robots must process real-time multimodal data and make decisions, meaning agents should also exhibit rapid response and multimodal processing capabilities.

\subsubsection{Power System}

Numerous mature applications already exist in electric power and energy systems based on Agent and Multi-Agent systems. For instance, \citet{kilkki2014agent} comprehensively reviews agent-based modeling and simulation applications in smart grids. The paper introduces the characteristics and challenges of smart grids. It classifies and compares agent-based modeling and simulation methods, discussing the application of agent-based models in different scenarios in smart grids. \citet{merabet2014applications} reviews MAS in smart grids, introducing the concept and characteristics of MAS and discussing the application scenarios, key technologies, and challenges of MAS in smart grids. \citet{ghazzali2020fixed} examines the fixed-time distributed voltage and reactive power compensation in islanded microgrids using sliding-mode and multi-agent consensus design methodologies. \citet{shinde2019agent} reviews the literature on agent-based modeling applications in various electricity markets. \citet{may2023multi} employs MARL to design dynamic pricing policies for energy markets under climate change scenarios.

LLM-based agent research in electric power and energy is developing, with relatively few related studies. Future research may explore the following directions:
\begin{itemize}

\item \textbf{Smart Grid Management and Optimization}: By emulating the behavior and interactions of power plants, transmission lines, and electricity-consuming equipment, LLM-based agents can effectively model challenges in smart grids. These challenges include power generation, transmission, distribution, and electricity consumption management. The assessment of grid stability, energy efficiency, and power dispatch is also possible with these agents.
\item \textbf{Distributed Energy Resource Scheduling}: The scheduling and optimization of distributed energy resources, such as solar energy, wind energy, and energy storage devices, can be modeled using LLM-based agents. These agents allow for examining distributed energy resources' power generation effects, market competition, and energy complementarity.
\item \textbf{Energy Market and Trading Mechanisms}: LLM-based agents are suitable for simulating issues in energy markets, such as supply-demand balance, price formation, and trading mechanisms. Precisely, they can simulate the behavior and interactions of energy producers, consumers, and trading platforms, analyzing energy markets' competitive landscape, price fluctuations, and trading efficiency.
\end{itemize}
Large-scale integration and collaborative optimization pose significant challenges in the context of renewable energy and distributed energy resources. To achieve efficient operation and sustainable development of power systems, LLM-based agents must consider various energy types, multi-level grid structures, and complex market environments. Moreover, developing related technical standards and specifications is necessary to facilitate the widespread application and promotion of multi-agent systems in smart grids. This will enhance the interoperability and scalability of these systems while reducing the difficulty and cost of integration. Through extensive research and innovation, LLM-based agents are anticipated to play a crucial role in smart grid management and optimization, distributed energy resource scheduling, and energy market and trading mechanisms, ultimately contributing to the sustainable development of power systems.


\subsubsection{Transportation System} Transportation has attracted extensive research interest in agent systems. MARL can be employed to coordinate multiple traffic signals to optimize traffic flow, reduce congestion, and enhance road traffic efficiency. \citet{zeng2018adaptive} introduces a method for controlling traffic signals utilizing deep Q-learning. \citet{chu2019multi} applies distributed MARL techniques to coordinate traffic signals in large-scale urban road networks to minimize traffic congestion.

Research on LLM-based agents in transportation is currently in its nascent stage. \citet{da2023llm} employs LLM to comprehend and analyze system dynamics through context-based prompts for reasoning. By leveraging the reasoning capabilities of LLM, it is possible to understand how weather conditions, traffic conditions, and road types affect traffic dynamics. Subsequently, the agent takes actions based on real-world dynamics and learns more realistic strategies accordingly. TrafficGPT \cite{zhang2023trafficgpt} combines LLM with traffic domain expertise to enhance traffic management effectiveness. Moreover, it equips LLM with the capability to visualize, analyze, and process traffic data, offering valuable decision support for urban transportation system management. DiLu \cite{wen2023dilu} integrates reasoning and reflection modules, allowing autonomous driving systems to make decisions grounded in common-sense knowledge.

LLM-based agents can investigate and contribute to the following aspects in the field of transportation: LLM-based agents can manage traffic signals, optimizing them according to real-time traffic flow and demand to reduce congestion and waiting times. Compared to traditional methods, dispatchers can adjust signal cycles through natural language. However, due to the involvement of multi-objective optimization and decision-making, significant challenges arise for LLM's reasoning and decision-making capabilities. On the other hand, LLM-based agents can be employed to simulate vehicle travel and road condition changes in the traffic flow process, assisting researchers in understanding the characteristics and factors influencing traffic flow. For instance, by simulating the behavior and interactions of vehicles, roads, and traffic signals, LLM-based agents can analyze traffic congestion, accidents, and efficiency issues, providing a higher degree of simulation compared to the original implementation, as LLM-based agent-simulated vehicles more closely resemble human decision-making.

For transportation systems, it is typically essential to effectively process real-time data and optimize decision-making based on real-time traffic flow and demand. LLM-based agents need to exhibit a rapid response speed. Furthermore, efficiently implementing traffic signal control and scheduling strategies when facing multiple optimization objectives and decision factors remains challenging.

\subsubsection{Industrial Control System} In the realm of Agent and Multi-Agent research, the current state of investigation regarding agent-based intelligent manufacturing systems is reviewed by \citet{shen1999agent}, with a particular emphasis on production scheduling and resource optimization concerns. \citet{shen2006applications} comprehensively examines agent-based system applications within the intelligent manufacturing domain.

At present, LLM-based agent applications in industrial control and engineering encompass works such as the study by \citet{xia2023towards}, which introduces an innovative framework integrating LLMs, digital twins, and industrial automation systems for the intelligent planning and control of production processes. The authors establish two categories of intelligent agents: a managerial agent functioning at the apex of the automation module, responsible for coordinating various module skills to devise production plans, and an operational agent situated within a specific automation module, orchestrating multiple functions to execute the provided skills. In energy-efficient lighting systems, \cite{nascimento2023gpt} employs sensors, actuators, and neural networks, achieving superior decision-making and adaptability by incorporating GPT-4 without requiring extensive training. In the field of chip design, an LLM-based agent is utilized by \citet{li:hal-04175312} to assist in the development of Finite-Difference Time-Domain (FDTD) simulation code and deep reinforcement learning code, culminating in optimized Photonic Crystal Surface Emitting Laser (PCSEL) structures for advanced silicon photonics and photonic integrated circuit applications.

The potential of LLM-based agents in industrial process control and optimization is promising, encompassing tasks such as data collection simulation, control strategy formulation, and equipment modification. LLM-based agents can assess industrial processes' stability, production efficiency, and energy consumption by emulating the behavior and interactions of sensors, controllers, and actuators. One challenge LLM-based agents face is bridging the gap between real-world task planning and text task planning, thereby augmenting their practical applicability in industrial process control and optimization. Another challenge pertains to addressing the complexities and scalability concerns that emerge from the multiple levels, roles, and objectives inherent in industrial process control and optimization.

\subsubsection{Medical System} Agent systems exhibit many applications within medical and pharmaceutical research, spanning areas such as drug discovery and optimization, exploration of drug mechanisms, and pharmacokinetic simulations. \citet{an2001agent} demonstrates the employment of agent-based computer simulations in biomedical research, including drug discovery and optimization processes. \citet{ekins2006pathway} presents agent-based pathway mapping tools for high-throughput data analysis, covering aspects such as drug mechanism investigation and drug target identification. \citet{walker2004epitheliome} proposes an agent-based model of cellular social behavior for simulating personalized drug treatments and precision medicine. \citet{singhal2022large} discusses the enhancement of LLMs within medical and clinical domains. \citet{zhavoronkov2019deep} develops Generative Tensorial Reinforcement Learning (GENTRL) to design novel small molecules, optimizing the synthesized compounds' feasibility, novelty, and bioactivity.

Currently, research on LLM-based agents in medical science remains relatively scarce. \citet{williams2023epidemic} introduces a novel individual model paradigm to tackle the challenge of incorporating human behavior into epidemic models, with agents exhibiting multi-wave epidemic patterns following the epidemic period, mirroring those observed in recent pandemics. \citet{lobentanzer2023platform} employs general and biomedical-specific knowledge to address the LLM hallucination issue and seamlessly integrates prevalent bioinformatics techniques, enhancing its practical applicability and reliability. \citet{mehandru2023large} presents a novel evaluation framework, termed "Artificial-intelligence Structured Clinical Examinations" ("AI-SCI"), for assessing the performance of LLM agents in real-world clinical tasks.

LLM-based agents hold significant potential within the fields of medical and pharmaceutical research, encompassing aspects such as:

\begin{itemize}
\item \textbf{Disease Transmission and Epidemiological Modeling}: Through the simulation of various agents' behavior and interactions in disease transmission, including infected, susceptible, and recovered individuals, as well as processes such as individual mobility, social behavior, and disease state alterations, investigators can obtain a deeper understanding of disease transmission dynamics and develop effective control strategies.
\item \textbf{Drug Discovery and Optimization}: LLM-based agents can be utilized to replicate the screening, optimization, and evaluation procedures in drug discovery, thereby aiding researchers in identifying novel drugs with specific effects and applications. Specifically, by simulating the behavior and interactions of drug molecules, target proteins, and biological processes, LLM agents can examine pharmaceuticals' structure-activity relationship, pharmacodynamics, and pharmacokinetics.
\end{itemize}
Nevertheless, this field involves numerous highly complex biological systems, and addressing these complexity issues while ensuring model accuracy remains a significant challenge.

\subsubsection{Military System}

Agents and Multi-Agent Systems (MAS) hold substantial potential in military research, particularly in aiding researchers in comprehending the intricacies and dynamics of military issues through simulation and emulation. \citet{ilachinski2004artificial, cil2010multi} introduce multi-agent-based war simulation methodologies, encompassing war simulation and tactical analyses, military intelligence, and decision support. \citet{sycara2006literature} reviews the advancements in teamwork models, including multi-agent-based military communication and command and control systems.

Currently, there is limited research on LLM-based MAS within the military domain. Future exploration could focus on war simulation and tactical analysis, where LLM-based MAS can be employed to simulate combat actions and tactical decision-making during warfare. This may involve simulating combat units, commanders, and terrain environments with multiple cooperating or opposing agents. Such simulations aid researchers in evaluating the strengths and weaknesses of various tactical plans and analyzing combat effectiveness, battlefield situations, and tactical advantages. Another area of interest is military intelligence and decision support: LLM-based agents can be utilized to implement military intelligence and decision support, thereby enhancing the accuracy and efficiency of command decisions. Precisely, agents can simulate intelligence collection, analysis, and decision-making to achieve real-time intelligence analysis, early warning, and strategic planning. LLM-based agents can leverage their robust generalization capabilities in different military scenarios for planning, analysis, and decision-making.

Nevertheless, military research often requires consideration of numerous factors, such as ensuring a highly realistic simulation environment, incorporating more accurate representations of battlefield terrain, weather conditions, combat unit performance, and multi-level (e.g., strategic, operational, tactical) and multi-domain (e.g., land, sea, air, space, cyber) factors. Effective collaboration with human decision-makers is crucial to accurately reflect the complexity and dynamics of combat actions and tactical decision-making. Simultaneously, legal and ethical issues must be addressed. As artificial intelligence technology becomes increasingly prevalent in the military, the importance of legal and ethical considerations grows.

\section{Discussion}\label{discussion}
\subsection{Trend}\label{trend}

\paragraph{Evaluation}
LLM-based agents have shown remarkable capabilities in various domains, including specified task-solving, cooperation, and human interaction. However, assessing their performance quantifiable and objectively remains a challenge. 
\begin{itemize}
    \item \textbf{Foundational Capabilities:} As the field of research on LLM-based agents continues to advance, the foundational competencies of these agents have reached a stage of relative stability, underscoring the imperative need for standardized assessments of these foundational capabilities. Notably, benchmarks such as Minecraft~ \cite{wang2023voyager,wang2023describe,zhu2023ghost} and Tachikuma~ \cite{liang2023tachikuma} have been introduced to gauge the understanding of LLM-based agents in comprehending complex problems and engaging in logical reasoning. Additionally, the AgentSims~ \cite{lin2023agentsims} is a versatile framework for evaluating the agent's planning and decision-making skills, including its ability to make informed decisions in various contexts. AgentBench~ \cite{liu2023agentbench} provides a comprehensive platform for assessing agents' foundational capabilities holistically. Evaluating tool and resource utilization by LLM-based agents has garnered considerable research attention. It is poised to evolve with the development of more standardized and finely-tuned assessment metrics and protocols in this domain. Notably, the ToolBench~ \cite{qin2023tool} and Gentopia~ \cite{xu2023gentopia}
    contribute to this assessment facet by ascertaining how agents can effectively utilize various tools and resources to accomplish tasks. Currently, retrieval capability is evaluated in online shopping scenarios such as WebShop~ \cite{yao2022webshop} and WebArena~ \cite{zhou2023webarena}. Information retrieval is essential for LLM-based agents to acquire updated knowledge, which should be included in the benchmark of tool utilization. Memory storage, retrieval, and memory form mechanisms are critical designs for LLM-based agents to maintain long-term contextual understanding and meaningful behavior. Quantified metrics and well-designed benchmarks in the memorization capability have been discussed in ~ \cite{zhong2023memorybank}, and enlarged tasks and metrics should be included to facilitate LLM-based agents' more anthropopathic memory behavior.
    
    \item \textbf{Domain-based Evaluation:} Assessing the performance of LLM-based agents requires benchmarking both the executive environment and the specified tasks. Simply relying on MBPP \cite{austin2021program} and HumanEval \cite{chen2021evaluating} benchmarks is not enough because LLM-based agents can observe the runtime execution results and perform code re-generation, such as MetaGPT~ \cite{hong2023metagpt} and ChatDev~ \cite{qian2023communicative}. Therefore, it is necessary to design task-level definition and evaluation protocols, as AgentBench shows. Furthermore, developing and announcing task-based benchmarks in diverse domains such as law and medicine are imperative to propel research and application of domain-specific LLM-based agents. Such benchmarks serve as critical reference points for assessing the efficacy and competence of these agents within specialized fields. Meanwhile, in psychology, assessment metrics such as the emotional assessment of LLM-based agents and treatment outcome assessment about applying LLM-based agents rely on human feedback and comparison as reported in ~\citet{huang2023emotionally}, datasets and evaluation mechanisms are essential. 
    
\end{itemize}

\paragraph{Continual Evolution} 
When operating in complex and dynamic environments, LLM-based agents typically require the ability to evolve continuously, adapting their parameters, memory, and objectives accordingly.

\begin{itemize}
    \item \textbf{Continual Learning and Self-training}: A crucial aspect of LLM-based agents is their capacity for continuous learning and adaptation. As tasks and domains evolve, agents must acquire new knowledge and skills without losing previously learned information. Techniques such as lifelong learning and meta-learning can enhance the agent's reasoning ability, enabling it to generalize and apply knowledge to novel situations. Furthermore, efficiently utilizing the agent's memory can improve its intrinsic generalization capabilities. Developing effective mechanisms for continual learning and self-training is essential for LLM-based agents' long-term success and applicability across various domains. Research in this area should focus on devising robust algorithms and models that allow agents to learn from diverse information sources, including textual data, user interactions, and real-world experiences. 
    \item \textbf{Self-Evaluation and Dynamic Goals}: LLM-based agents should possess self-evaluation and goal-setting capabilities to enhance performance and adapt to changing environments. It is crucial for agents to assess the feedback from their environment and comprehend any criticism of their behavior.  LLM-based agents can effectively learn from feedback and extract and retain critical experiences utilizing more efficient programming mechanisms.  LLM-based agents can evaluate signals or quantitative metrics and qualitative feedback, enhancing their ability to process evaluations. This ability involves assessing their strengths and weaknesses, identifying areas for improvement, and setting realistic self-improvement goals. Agents should also monitor their progress toward these goals and make necessary adjustments to remain on track. Developing self-evaluation and dynamic goal-setting mechanisms will enable LLM-based agents to become more autonomous and adaptive, leading to improved performance and more effective human-agent collaboration.
    \item \textbf{Adaptability}: The success of LLM-based agents critically depends on their ability to adapt to new environments, tasks, and user preferences. This adaptability encompasses several aspects, including understanding and adjusting to user needs, adapting to different communication styles, and swiftly learning new tasks and domains. Research in this area should concentrate on creating models and algorithms that enable agents to learn from their experiences and interactions with users, allowing them to adjust their behavior and strategies accordingly. Developing robust LLM and Rethink techniques will also enable LLM-based agents to apply their knowledge and skills in novel situations, ultimately leading to more versatile and effective agents.
\end{itemize}

\paragraph{Enhancement of Multimodal Capabilities}

Agents must manage multimodal information in real-world situations, encompassing images, videos, and speech. The incorporation of additional multimodal models can equip LLM-based agents with multimodal proficiencies. This process typically entails converting multimodal input into textual data, utilizing LLMs for inference and planning, and employing multimodal models for output generation. For example, MM-React \cite{yang2023mmreact} integrates ChatGPT with a visual expert pool to accomplish multimodal inference and action. IdealGPT \cite{you2023idealgpt} is a framework for the iterative decomposition of visual reasoning, employing LLMs to generate sub-questions, multimodal models to supply corresponding sub-answers, and LLMs to deduce the final response. \citet{di2023towards} propose a framework that amalgamates an RL-based agent trained from scratch with the advanced capabilities of LLMs and multimodal models. The agent can explain its multimodal environment, tasks, and actions through language. TaPA \cite{wu2023embodied} generates executable plans by aligning LLMs and visual perception models for real-world scenarios with physical scene constraints. ViperGPT \cite{surís2023vipergpt} combines visual and language models using code generation models to produce results for any query.

Conversely, recent large multimodal models (LMMs), such as GPT4-V \cite{yang2023dawn}, miniGPT-v2 \cite{chen2023minigptv2}, LLaVA \cite{liu2023visual}, and PALM-E \cite{driess2023palme}, have exhibited robust image content comprehension capabilities. In the future, when constructing agents with LMMs, it will no longer be necessary to convert images into text before inputting them into LLMs. Instead, LMMs can directly execute multimodal task planning and reconsideration based on the current image input, enhancing information utilization efficiency and multimodal task processing performance.

\subsection{Challenges}\label{challenge}

\subsubsection{Intrinsic Constraints of LLMs}
LLMs provide the foundation for LLM-based agents, facilitating planning and reconsideration capacities, natural language expression, and robust generalization across diverse tasks. Nevertheless, LLMs often face constraints due to context length \cite{zhao2023survey}, which may result in the loss of essential information when processing extensive articles or complex dialogues \cite{liu2023lost}. Another concern is the generation of invalid data and hallucinations by the LLM \cite{zhang2023hallucination}. Despite the ability of LLMs to produce fluent and ostensibly plausible text, they may generate irrelevant, invalid, or even erroneous information. This phenomenon arises from LLMs acquiring extraneous data or incorrect patterns during training. Such issues considerably influence the efficacy of LLMs, subsequently impacting the overall performance of both LLM-based agents and LLM-based MAS.

\subsubsection{Dynamic Scaling}
As the deployment of LLM-based MAS becomes more widespread, the system must be capable of dynamic expansion across various hardware and software environments, adjusting its scale and performance in response to demand. However, the implementation of dynamic scaling presents several challenges, including:

\begin{itemize}
\item \textbf{Adaptability}: The system must be capable of adjusting its scale and performance to meet diverse task requirements and computing environments. This requires robust adaptive capabilities, including automatic adjustments to the number of agents, the sizes of various memory spaces, and conversion strategies. Researchers may employ adaptive algorithms, such as reinforcement learning and genetic algorithms, for automatic optimization and adjustment to achieve this adaptability.
\item \textbf{Resource allocation and management}: Dynamic scaling requires the adaptive expansion of computing and storage resources for MAS. In the case of CPDE MAS (Section \ref{CPDE}), where a single LLM is responsible for role allocation and action planning, dynamic scaling must consider the LLM's allocation and planning relative to a varying number of agents and the resource consumption of LLM's reasoning. For instance, \citet{yue2023large} explores the construction of LLM cascades to reduce the cost of employing LLMs, particularly in executing reasoning tasks.
\end{itemize}

\subsubsection{Security and Trust}

The allocation of appropriate permissions and the assurance of system security are critical for LLM-based agents \cite{schwartz2023enhancing}. Given that these agents can exchange information and resources, excessive permissions could lead to incorrect decisions and actions, impacting overall system performance and raising security issues. How can we prevent harmful errors, thus preserving the hard-won trust of humans and enterprises? To address this issue, developing an effective permission allocation mechanism that fosters efficient collaboration among distinct agents without exceeding their designated authority is imperative. Additionally, the importance of conducting reliability tests cannot be overstated. For instance, ToolEmu \cite{ruan2023identifying} utilizes LLM to simulate tool execution, showcasing its ability to evaluate LLM-based agents across various tools and scenarios. This approach enables the detection of agent failures and the quantification of associated risks.

\section{Conclusion}

This paper comprehensively reviews LLM-based agents' current research status, applications, and prospects. It begins by tracing the development from agents to RL-based agents and subsequently to LLM-based agents, followed by an introduction to the fundamental concepts of LLM-based agents, including their definition, planning capabilities, memory, rethinking capabilities, action, and external environment. Subsequently, the paper elaborates on the multi-role relationships, planning types, and enhanced communication methods of LLM-based MAS. Additionally, it discusses the potential development prospects and challenges of LLM-based agents in various fields and proposes possible solutions. Lastly, the paper delves into the development trends and challenges LLM-based agents face, such as LLM's inherent limitations, the dynamic expansion of MAS, and security and trust issues. Although current research is still far from achieving AGI, we believe that LLM-based agents can represent a significant step forward.

\bibliography{references}

\end{document}